\documentclass[webpdf,contemporary,large]{oup-authoring-template}%
\makeatletter
\def\@DOI{}        
\def\@doi{}        
\makeatother

\renewcommand{\paragraph}[1]{\smallskip\noindent{\bf #1}}
\usepackage{multirow}
\usepackage{natbib}
\usepackage{wrapfig}

\DeclareMathAlphabet\mathbfcal{OMS}{cmsy}{b}{n} 

\usepackage{stackrel,amssymb,amsmath}
\usepackage{graphicx}
\usepackage[margin=1in]{geometry}
\usepackage{xcolor}
\definecolor{darkgreen}{RGB}{34, 139, 34} 
\definecolor{ipcolor}{RGB}{242,170,60}
\definecolor{bpcolor}{RGB}{55,126,247}
\usepackage{xspace}
\usepackage{mathtools}
\usepackage{hyperref}  
\hypersetup{
    colorlinks=true,
    linkcolor=blue, 
    urlcolor=blue,  
    citecolor=blue, 
}

\renewcommand{\cite}[1]{\citep{{#1}}} \newcommand{\namecite}[1]{\citet{{#1}}} 

\usepackage{verbatim}

\usepackage{courier}
\usepackage{float}
\usepackage{tikz}
\usetikzlibrary{positioning, arrows.meta}
\usetikzlibrary{shapes,snakes}
\usepackage[capitalise]{cleveref}
\usepackage{tablefootnote}

\newcommand{\LM}[1]{\ensuremath{p_{#1}}\xspace}

\renewcommand{\vec}[1]{\ensuremath{\boldsymbol{{#1}}}\xspace}

\newcommand{\nts}{\ensuremath{\text{\it nt}}\xspace}

\newcommand{\pairs}{\ensuremath{\mathit{pairs}}\xspace}

\newcommand{\unpaired}{\ensuremath{\mathit{unpaired}}\xspace}

\newcommand{\notes}[1]{}

\newcommand{\ith}[1]{\ensuremath{i^{{th}}}}

\newcount\permx
\newcount\permy
\def\permdot#1#2{
\permx=#1 \advance\permx by-1
\permy=#2 \advance\permy by-1
\psframe[fillcolor=black, fillstyle=solid]
(\permx,\permy)(#1, #2)
}

\newcommand{\argmax}{\operatornamewithlimits{\mathrm{argmax}}}

\newcommand{\argmin}{\operatornamewithlimits{\mathrm{argmin}}}

\newcommand{\vecx}{\ensuremath{\vec{x}}\xspace}
\newcommand{\vecy}{\ensuremath{\vec{y}}\xspace}

\newcommand{\vecystar}{\ensuremath{\vecy^{\!\!\;\text{$\star$}}}\xspace}

\newcommand{\vecxstar}{\ensuremath{\vecx^\star}\xspace}

\newcommand{\smallnt}[1]{\ensuremath{_{\mbox{\tiny PP}}}\xspace}

\iffalse

\else

\fi

\newcommand{\defeq}{\ensuremath{\stackrel{\Delta}{=}}\xspace}

\renewcommand{\to}{\ensuremath{\!\rightarrow\!}\xspace}

\newcommand{\smallurl}[1]{{\scriptsize \url{#1}}}

\newcommand{\nucA}{\ensuremath{\text{\tt A}}}
\newcommand{\nucU}{\ensuremath{\text{\tt U}}}
\newcommand{\nucC}{\ensuremath{\text{\tt C}}}
\newcommand{\nucG}{\ensuremath{\text{\tt G}}}

\newcommand{\freeenergy}{\ensuremath{\Delta\:\!\! G^{\scriptsize\circ\!}}\xspace}
\newcommand{\DeltaG}{\freeenergy}

\newcommand{\MFE}{\ensuremath{\text{MFE}}\xspace}
\newcommand{\MFEs}{\ensuremath{\text{MFEs}}\xspace}

\newcommand{\UMFE}{\ensuremath{\text{uMFE}}\xspace}

\newcommand{\ED}{\ensuremath{\text{ED}}\xspace}

\definecolor{intnull}{RGB}{213,229,255}
\definecolor{inteins}{RGB}{128,179,255}
\definecolor{intvier}{RGB}{42,127,255}
\definecolor{intdrei}{RGB}{0,85,212}
\definecolor{intvier}{RGB}{0,51,128}
\definecolor{intfunf}{RGB}{0,34,85}

\newsavebox\CBox

\definecolor{pair_color}{rgb}{0.0,0.0,1.0}
\definecolor{mismatch_color}{rgb}{1.0,0.0,0.0}
\definecolor{trimismatch_color}{rgb}{0.71,0.4,0.11}

\definecolor{designable}{rgb}{0.0,0.0,0.0}
\definecolor{undesignable}{rgb}{0.71,0.4,0.11}
\definecolor{unknown}{rgb}{0.5,0.5,0.5}

\newcommand{\LSL}{\ensuremath{\mathcal{L}_{\text{SL}}}\xspace}
\newcommand{\RfamlearnMFE}{Rfam-Learn-MFE\xspace}
\newcommand{\Eternaweb}{Eterna-web\xspace}
\newcommand{\RandomMFE}{Random-MFE\xspace}

\newcommand{\Ytest}{\ensuremath{\mathbfcal{Y}_{\text{test}}}\xspace}
\newcommand{\Ytrain}{\ensuremath{\mathbfcal{Y}_{\text{train}}}\xspace}
\newcommand{\Ytrainraw}{\ensuremath{\mathbfcal{Y}_{\text{train\_raw}}}\xspace}
\newcommand{\YXtrain}{\ensuremath{\mathbfcal{YX}_{\text{train}}}\xspace}

\newcommand{\YRLraw}{\ensuremath{\mathbfcal{Y}_{\text{RL\_raw}}}\xspace}
\newcommand{\YRLlarge}{\ensuremath{\mathbfcal{Y}_{\text{RL\_large}}}\xspace}

\newcommand{\YRL}{\ensuremath{\mathbfcal{Y}_{\text{RL}}}\xspace}

\newcommand{\dstruct}{\ensuremath{d_{\text{struct}}}\xspace}

\newcommand{\leftrarrows}{\mathrel{\raise.75ex\hbox{\oalign{%
  $\scriptstyle\leftarrow$\cr
  \vrule width0pt height.5ex$\hfil\scriptstyle\relbar$\cr}}}}
\newcommand{\lrightarrows}{\mathrel{\raise.75ex\hbox{\oalign{%
  $\scriptstyle\relbar$\hfil\cr
  $\scriptstyle\vrule width0pt height.5ex\smash\rightarrow$\cr}}}}
\newcommand{\Rrelbar}{\mathrel{\raise.75ex\hbox{\oalign{%
  $\scriptstyle\relbar$\cr
  \vrule width0pt height.5ex$\scriptstyle\relbar$}}}}

\makeatletter
\def\leftrightarrowsfill@{\arrowfill@\leftrarrows\Rrelbar\lrightarrows}
\newcommand{\xleftrightarrows}[2][]{\ext@arrow 3399\leftrightarrowsfill@{#1}{#2}}
\makeatother

\makeatletter

\def\ps@firstpage{%
  \let\@oddhead\@empty
  \let\@evenhead\@empty
  \def\@oddfoot{\hfil\thepage\hfil}%
  \def\@evenfoot{\hfil\thepage\hfil}%
}

\let\old@maketitle\maketitle
\renewcommand{\maketitle}{%
  \begingroup
    \let\thanks\@gobble
    \let\footnotetext\@gobble   
    \let\@footnotetext\@gobble  
    \old@maketitle
  \endgroup
  \thispagestyle{firstpage}%
}

\makeatother

\begin{document}

\journaltitle{}
\copyrightyear{}
\pubyear{}
\access{}
\appnotes{Paper}

\title{Designing RNAs with Language Models}

%
%
\author[1$\diamond$]{\large Milan Gautam} 
\author[1$\diamond$]{\large Ning Dai} 
\author[1]{\large Tianshuo Zhou} 
\author[1]{\large Bowen Xie} 
\author[3]{\large David Mathews} 
\author[1,$\ast$]{\hspace{-0.2cm}\large Liang Huang} 

\authormark{Gautam and Dai et al.}
%
\address[1]{\orgdiv{School of EECS}}
\address[2]{\orgdiv{Dept.~of Biochemistry \& Biophysics}, \orgname{Oregon State University}, \country{USA}}
\address[3]{\orgdiv{Dept.~of Biochemistry \& Biophysics}}
\address[4]{\orgdiv{Center for RNA Biology}}

\address[5]{\orgdiv{Dept.~of Biostatistics and Computational Biology}, \orgname{University of Rochester Medical Center}, \country{USA}}

\abstract{RNA design, the task of finding a sequence that folds into a target secondary structure, has broad biological and biomedical impact but remains computationally challenging due to the exponentially large sequence space and exponentially many competing folds. Traditional approaches treat it as an optimization problem, relying on per-instance 
heuristics or constraint-based search. We instead reframe RNA design as conditional sequence generation and introduce a reusable neural approximator, instantiated as an autoregressive language model (LM), that maps target structures directly to sequences. We first train our model in a supervised setting on random-induced structure-sequence pairs, and then use reinforcement learning (RL) to optimize end-to-end metrics. We also propose methods to select a small subset for RL that greatly improves RL efficiency and quality. Across four datasets, our approach outperforms 
state-of-the-art systems on key metrics such as Boltzmann probability while being 1.7$\times$ faster, establishing conditional LM generation as a scalable, task-agnostic alternative to per-instance optimization for RNA design.
Our code and data are available at
{\small \textbf{\url{https://github.com/KuNyaa/RNA-Design-LM}}}.}


\maketitle

\section{Introduction}
\label{sec:intro}

RNA plays a central role in cellular processes such as transcription, translation, catalysis, and gene regulation~\cite{eddy:2001,doudna+cech:2002,bachellerie+:2002}. 
Its biological and biomedical importance is underscored by RNA viruses like SARS-CoV-2, as well as the Nobel Prizes in 
both 2023 (mRNA vaccine) and 2024 (microRNAs).
Since structure determines function, designing artificial RNA molecules with specific secondary structures enables a wide range of applications, including artificial ribozymes~\cite{dotu+:2014,yamagami+:2018}, microRNAs~\cite{schwab+:2006}, aptamers~\cite{hamada:2018}, and riboswitches~\cite{bauer+suess:2006,findeiss+:2017}.  

The problem of \emph{RNA design}, also known as {\em RNA inverse  folding}, aims to find an RNA sequence that folds into a given target structure~\cite{hofacker+:1994,andronescu+:2004,garcia+:2013rnaifold,zadeh+:2011nupack,zhou+:2023samfeo}. 
Computationally, this is an exceptionally challenging problem due to two levels of combinatorial explosion,
both in the {\em sequence (design) space} and {\em structure (folding) space}.  
For any target structure, there are exponentially many candidate designs,
while for each design, there are also exponentially many alternative structures that compete with the target.  
Indeed, the problem has been proven NP-hard under simplified energy models~\cite{bonnet+:2020}.  

Existing approaches to RNA design 
can be broadly categorized into three camps.
The most common one is heuristic local search that iteratively mutates candidate sequences, such as RNAinverse~\cite{hofacker+:1994}, NEMO~\cite{portela:2018nemo} and SAMFEO~\cite{zhou+:2023samfeo}.
Operating on one or a few sequences at a time,
they cannot keep up with the exponential growth of the design space and struggle at long and hard-to-design structures.
To address this fundamental problem,
the second camp recursively decomposes large structures
into smaller ones,
which includes
RNA-SSD~\cite{andronescu+:2004}, RNAiFold~\cite{garcia+:2013rnaifold},
and NUPACK~\cite{zadeh+:2011nupack}.
Nevertheless, these methods are computationally expensive, 
and generally underperform the first camp.
A third group uses  
reinforcement learning (RL), formulating RNA design as a sequential decision-making problem~\cite{eastman+:2018,runge+:2019,obonyo+:2022}.  
However, the performance of 
 these vanilla RL systems trained from scratch
lag behind strong heuristic designers.  

In this work, we propose a fundamentally different perspective: RNA design as \emph{conditional sequence generation}.  
Instead of solving a combinatorial optimization problem from scratch for each structure, we learn a reusable (shared) \emph{neural approximator solver}.  
We make the following contributions:
\begin{enumerate}
\item We formulate RNA design as conditional sequence generation using an autoregressive language model.
\item We devise a constrained decoding algorithm to ensure the validity of  designs for the input structure.
\item We develop a unified framework combining pretraining, supervised learning (SL) on 
structure–sequence pairs,  
and reinforcement learning (RL) 
on selected structures to directly optimize folding success.  
\item We show that, surprisingly, SL on random-induced structure-sequence pairs unrelated to the test data yield competitive results on the latter.
\item We develop methods to select a small subset of structures for efficient RL based on sample diversity.
\item We utilize
 fast parallel sampling of large amounts of designs on modern GPUs,
 and our best model (SL+RL) outperforms the state-of-the-art systems in key metrics on four test sets, while being 1.7$\times$ faster.
\end{enumerate}

\vspace{-0.7cm}
\section{Preliminaries: RNA Folding and Design}
\label{sec:prelim}


\subsection{RNA Sequences and Structures}

An RNA sequence $\vecx$ of is a string of nucleotides 
$x_1 x_2 \dots x_n$, where $x_i \in \mathcal{N}$ and 
$\mathcal{N} \defeq \{\nucA, \nucC, \nucG, \nucU\}$.  
A pseudoknot-free secondary structure $\vecy$ is represented as a dot-bracket string of the same length.  
Each character $y_i \in \{ \texttt{.}, \texttt{(}, \texttt{)}\}$ indicates whether nucleotide $x_i$ is unpaired, paired downstream, or paired upstream, respectively.  
Base pairs are restricted to Watson–Crick–Franklin or wobble pairs:  
$\mathcal{P} = \{ \texttt{CG}, \texttt{GC}, \texttt{AU}, \texttt{UA}, \texttt{GU}, \texttt{UG}\}$.  
For a given sequence $\vecx$, its \emph{ensemble} $\mathcal{Y}(\vecx)$ consists of all possible secondary structures.  
In thermodynamic RNA folding models, \emph{Gibbs free energy change} $\DeltaG(\vecx, \vecy)$ is used to characterize the stability of secondary structure $\vecy \in \mathcal{Y}(\vecx)$. The lower the free energy~$\DeltaG(\vecx, \vecy)$, the more stable the structure $\vecy$ for $\vecx$. The structure with the \emph{minimum free energy} is the most stable one in the ensemble, i.e., 
the {\em \text{MFE} structure},
\vspace{-5pt}
\begin{equation}
\text{MFE}(\vecx) \defeq \argmin_{\vecy \in \mathcal{Y} (\vecx)} \DeltaG(\vecx, \vecy).
\vspace{-5pt}
\end{equation}

Note that for most methods for secondary structure prediction, ties for $\argmin$ are broken arbitrarily when there are multiple lowest free energy structures.
This issue was often neglected in the literature, but it deserves clarification here. To be precise, we define 
\(\MFEs(\vecx) \defeq \{\vecy \mid 
\DeltaG(\vecx, \vecy) = \min_{\vecy' \in \mathcal{Y}(\vecx)} \DeltaG(\vecx, \vecy')\}\)
to be the {\em set of MFE structures} for $\vecx$. When 
$|\MFEs(\vecx)|=1$,
we say $\vecx$ has a {\em unique MFE} (uMFE) structure. 

The \emph{partition function} 
\(
Q(\vecx) \stackrel{\Delta}{=}  \sum_{\vecy \in \mathcal{Y}(\vecx)} e^{-\DeltaG(\vecx, \vecy)/RT}
\)
sums the contribution of all structures in the ensemble,
where $R$ is the molar gas constant and $T$ is the absolute temperature. Accordingly, the Boltzmann 
probability of a sequence $\vecx$ folding into a structure $\vecy$ is defined as
\(
p(\vecy \mid \vecx) = \frac{e^{-\DeltaG(\vecx, \vecy)/RT}}{Q(\vecx)}.
\)

\vspace{-0.2cm}
\subsection{RNA Design Problem}

Given a target structure~\vecystar, RNA design (inverse folding problem; Fig.~\ref{fig:folding-design}) aims to find a suitable RNA sequence~\vecx 
that can naturally and easily fold into $\vecystar$,
within the design space $\mathcal{X}(\vecystar)$ 
of all valid sequences for $\vecystar$:
\vspace{-3pt}
\begin{equation}
\mathcal{X}(\vecystar) \defeq \{ \vecx \in \mathcal{N}^{|\vecystar|} \mid \forall (i,j)\in \pairs(\vecystar), x_i x_j \in \mathcal{P}\}
\end{equation}
We formulate the RNA design problem as optimizing  
a scoring function
$f(\vecx, \vecystar)$ 
over the design space $\mathcal{X}(\vecystar)$:

\vspace{-10pt}
\begin{equation}
\vecxstar = \argmax_{\vecx \in \mathcal{X}(\vecystar)} f(\vecx, \vecystar)
\end{equation}
where
the scoring function quantifies how ``easily'' 
$\vecx$ folds into $\vecystar$. 
In practice, RNA design quality
is evaluated 
using several standard metrics
\citep{anderson+:2016,zhou+:2023samfeo} which we adopt below.
The first two (ensemble-based)
are  more important than the last two (MFE-based),
because even if \vecx is an MFE solution there can still be many  structures with similar energies,
resulting in arbitrarily low probability of \vecystar in ensemble.

\vspace{-5pt}
\begin{itemize}
    \item \textbf{Boltzmann probability} \( p({\vecystar} \mid \vecx) \) meansures
 how likely \( \vecx \) folds into 
    \( {\vecystar} \) among all possible structures.

    \item \textbf{Normalized Ensemble Defect} \( \text{NED}(\vecx, {\vecystar}) \)
     quantifies how well the ensemble of structures formed by \( \vecx \) 
    aligns with the target, penalizing competing folds~\cite{zadeh+:2011}:
 \(       \text{NED}(\vecx, {\vecystar}) 
        = \frac{1}{|\vecx|} \sum_{\vecy \in \mathcal{Y}(\vecx)} 
        p(\vecy \mid \vecx)\, \dstruct({\vecystar}, \vecy),
  \) 
    where 
    $\dstruct(\vecy, \vecy') \stackrel{\Delta}{=} 
        |\vecy| 
        - 2 \cdot |\pairs(\vecy) \cap \pairs(\vecy')|
        - |\unpaired(\vecy) \cap \unpaired(\vecy')|$ is the structural distance.
   \item {\bf MFE solution}: $f(\vecx,\vecystar)=1$ iff.~$\vecystar\in \MFEs(\vecx)$. 
   \item {\bf \UMFE solution}: $f(\vecx,\vecystar)\!=\!1$ iff.~$\MFEs(\vecx)=\{\vecystar\}$. 

\begin{figure}[t]
\hspace{-0.4cm}
\begin{tabular}{ccc}
\begin{tabular}{c}
\includegraphics[width=0.2\linewidth]{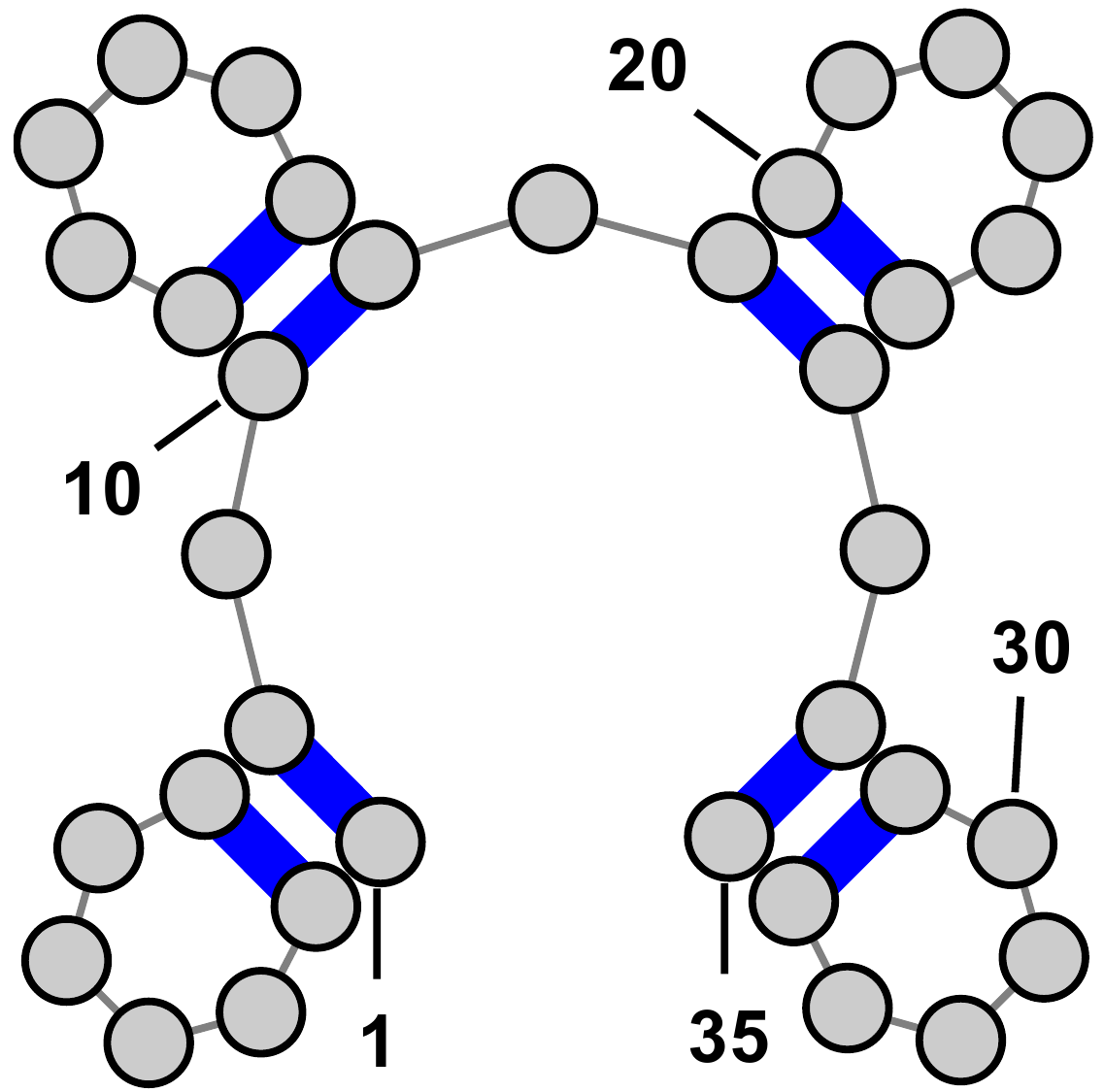}
\end{tabular}
\hspace{-1.1cm}\vecystar\hspace{.6cm}
&
\hspace{-0.5cm}
\(\xleftrightarrows[\text{design}]{\text{folding}}\)
\hspace{-0.3cm}
& \scriptsize \vecx: \texttt{GGGAAACCAGGGAAACCAGCGAAAGCACCUACGGG}
\end{tabular}
\caption{RNA design is the inverse problem of RNA folding.\label{fig:folding-design}}
\end{figure}


\end{itemize}

\vspace{-0.5cm}
\section{Models and Constrained Decoding}
\label{sec:methods}


\subsection{RNA Design as Conditional Seq.~Generation}

We frame RNA design as a conditional sequence generation task.  
Given a target secondary structure~$\vecystar$, 
we use an autoregressive language model $\LM{\theta}(\cdot\mid\vecystar)$ parameterized by $\theta$ that defines the conditional distribution
\vspace{-8pt}
\[
p_\theta(\boldsymbol{x} \mid \vecystar) 
= \prod_{t=1}^{n} p_\theta(x_t \mid \vecx_{<t}, \vecystar),
\]
where $x_t \in \mathcal{N}$ is the nucleotide at position $t$ and $\vecx_{<t}$ is the sequence prefix.  
This autoregressive factorization enables the model to condition each nucleotide decision on both the target structure and the previously generated prefix, capturing local dependencies that influence valid folding.

\paragraph{Vocabulary and input.}
We employ a compact, domain-specific vocabulary designed for structure-conditioned RNA generation.  
It consists of the four nucleotides (\nucA, \nucC, \nucG, \nucU), the three dot–bracket structure symbols (\texttt{(}, \texttt{)}, \texttt{.}), and a small set of control tokens.  
The tokens \texttt{<struct>} and \texttt{</struct>} delimit the target structure, while \texttt{<bos>} and \texttt{<eos>} mark sequence boundaries.  
During inference, the structure is provided as a prefix prompt:
\vspace{-7pt}
\[
\texttt{<struct> ( . ( . . . ) ) </struct> <bos>}\vspace{-7pt}
\]
after which the model autoregressively emits nucleotides until producing an \texttt{<eos>} token:
\vspace{-7pt}
\[
\texttt{G A C U U A G C <eos>}.
\]

\subsection{Constrained Decoding}
\label{sec:decoding}

Autoregressive generation with a language model does not, by default, respect the biochemical pairing rules required for a sequence to be compatible with a given RNA secondary structure.  
When decoding nucleotides token by token, the model may freely emit bases that violate Watson--Crick--Franklin or wobble pairing at positions that correspond to paired sites in $\vecystar$, resulting in invalid sequences outside the valid design space $\mathcal{X}(\vecystar)$. 
To mitigate this mismatch between vanilla language model generation and RNA thermodynamic constraints, we employ a \emph{constrained decoding} mechanism that enforces base-pairing rules during generation.

In the aspect of formal language theory, our valid design space $\mathcal{X}(\vecystar)$ is a {\bf context-free language} (with long-distance dependencies) and needs a runtime stack to ensure valid decoding. However, our method below uses a (stack-based) precomputation to avoid the runtime stack, reducing the overhead.

\paragraph{Constraint rule.}
Let $\vecystar$ denote a dot–bracket structure.
We precompute a mapping $m(t)$ that maps the index of each 
``\texttt{)}'' to the index of its matching ``\texttt{(}''.  
At decoding step~$t$, the set of admissible nucleotides is defined as:
\vspace{-2pt}
\[
\mathcal{A}_t =
\begin{cases}
\{\nucA, \nucC, \nucG, \nucU\}, & y^\star_t\!\in\!\{\texttt{.},\texttt{(}\},\\
\mathrm{Comp}(x_{m(t)}), & y^\star_t\!=\!\texttt{)}
\end{cases}
\]
where $\mathrm{Comp}(\cdot)$ lists the nucleotides compatible with pairing constraints, i.e.,
\(\mathrm{Comp}(\nucA)=\{\nucU\}\),
\(\mathrm{Comp}(\nucC)=\{\nucG\}\),
\(\mathrm{Comp}(\nucG)=\{\nucC, \nucU\}\), and
\(\mathrm{Comp}(\nucU)=\{\nucA, \nucG\}\).

\paragraph{Implementation.}
At each decoding step, we mask the logits of all invalid nucleotides according to $\mathcal{A}_t$ and renormalize:
\(
\tilde{p}_\theta(x_t \mid \vecx_{<t},\vecystar)
\propto 
\mathbf{1}[x_t\!\in\!\mathcal{A}_t]\,
p_\theta(x_t \mid \vecx_{<t},\vecystar).
\)
A lightweight callback computes $\mathcal{A}_t$ on-the-fly from $y^\star_t$ and the nucleotide previously generated at its paired site.  
After position $n$ has been filled, only \texttt{<eos>} remains valid, ensuring that the generated sequence length matches the target structure.

\paragraph{Effect.}
This decoding strategy guarantees structural validity by construction and reduces the effective search space from $4^n$ (regular language $\{\nucA,\nucC,\nucG,\nucU\}^n$) to $6^m 4^u$ (context-free language $\mathcal{X}(\vecystar)$) for $m$ base pairs and $u$ unpaired positions.  
Its computational overhead is about 30\% (see Fig.~\ref{fig:combined_metrics_vs_length}), yet it substantially improves sample efficiency and training stability in both supervised and reinforcement learning settings.

\vspace{-0.2cm}
\section{Language Model Training}
\label{sec:training}
\subsection{Pretrained Model}

Large language models trained on massive, diverse text corpora have been shown to acquire broad inductive biases that transfer surprisingly well beyond natural language~\cite{touvron+:2023,yang+:2024}.  
Even relatively small decoder-only LMs develop stable attention patterns, long-range dependency modeling, and useful parameter initializations that make fine-tuning in downstream tasks significantly more data-efficient than training a transformer from scratch.  
Motivated by these observations, we begin with the Qwen2.5--0.5B~\cite{yang+:2024} pretrained model 
(GPT-style decoder-only Transformer)
as the backbone for our RNA design system.  
Although the original training domain is textual, our experiments show that the pretrained backbone provides substantially better optimization stability and sample efficiency than a randomly initialized transformer of identical architecture.  
In ablations, the scratch-trained model converges more slowly and yields lower folding performance, confirming that the pretrained initialization is a critical component of our approach.

\begin{figure}[h] 
    \centering
    \includegraphics[width=0.9\linewidth]{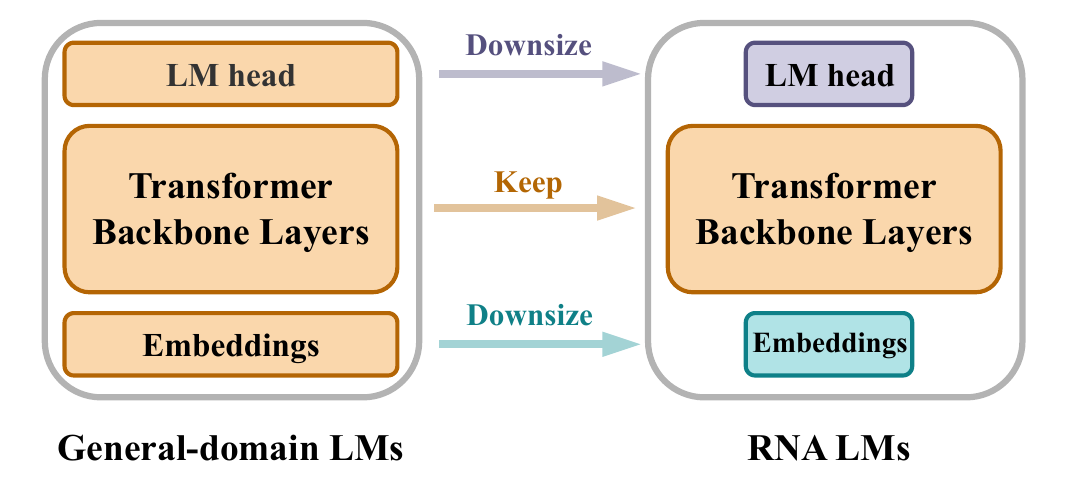}
	\vspace{-8pt}
    \caption{We convert a general-domain LLM into an RNA designer by keeping the pretrained transformer backbone  and shrinking the input and output layers. The original embedding and LM head are downsized and reinitialized to support RNA tokens.}
    \label{fig:RNA_LM_surgery}
\end{figure}

\paragraph{Surgery for RNA adaptation.}
To turn the general-domain LLM into an RNA-capable generator, we perform a minimal architectural modification. As illustrated in Fig.~\ref{fig:RNA_LM_surgery}, we keep all transformer backbone layers intact and modify only the input and output layers. The original embedding matrix and LM head are removed, and we introduce downsized, RNA-specific replacements built around the vocabulary defined above. This preserves the pretrained backbone while enabling the model to process RNA tokens and structural prompts.

\vspace{-15pt}

\paragraph{Why not use existing RNA LMs?}
Several RNA-focused language models have been proposed, but they are not well suited to our setting (see also Sec.~\ref{sec:related}):

\begin{enumerate}
    \item \textbf{Lack of structural input.}  
    Existing RNA LMs typically operate only on nucleotide sequences and do not accept dot--bracket structures as input, preventing direct structure-conditioned generation for design.

    \item \textbf{Architectural mismatch.}  
     Most publicly available RNA LMs are BERT-style encoders 
     rather than autoregressive decoders.  
    A GPT-style RNA LM has been reported, but no usable checkpoint has been released \cite{RNA-GPT}.
	
	\item \textbf{Tokenization mismatch.}  
    Existing RNA LMs adopt subword or BPE-style tokenization (e.g., merging adjacent nucleotides into one token, as in some recent generative RNA LMs~\cite{GenerRNA}).  
    This breaks the one-nucleotide-per-token setting that our constrained decoding relies on, making it much harder to control the validity of the decoded sequence: base-pairing constraints can no longer be enforced easily.

    \item \textbf{Software compatibility.}  
    Many RNA LMs are implemented in standalone codebases or lack Hugging-Face compatibility, making integration with our constrained decoding and RL pipeline difficult.
\end{enumerate}

\paragraph{Why not  encoder-decoder models?}
Although RNA design can be formulated as encoder-decoder translation ($\vecystar \to \vecx$), we adopt a decoder-only LM for high-throughput sampling and pretraining compatibility.
Encoder-decoder models need to compute cross-attention at each decoding step~\cite{vaswani+:2017}, which often dominates incremental decoding cost, limiting sampling throughput when generating many candidates per target~\cite{lu+:2024} (we use $10^4$ samples).
Moreover, conditioning via prefix prompting keeps supervised training and future RNA pretraining straightforward, and naturally supports mixing RNA sequences with sequence-structure pairs, whereas encoder-decoder pretraining on sequence-only data is not straightforward and requires denoising objectives or back-translation pipelines~\cite{raffel+:2020,lewis+:2020,sennrich+:2016}.

\subsection{Supervised Learning (SL)}

To initialize the model, we construct a training corpus of structure–sequence pairs.  
Following the knowledge distillation paradigm, we employ a strong optimization-based solver, SAMFEO~\cite{zhou+:2023samfeo}, to generate sequences for diverse target structures (see Sec.~\ref{sec:SLdata}).  
The resulting dataset
\(\YXtrain\) 
contains $\{(\vecystar, \vecx)\}$ pairs
where each $\vecx$ is a valid design for  $\vecystar$.  
The model is  trained by maximum likelihood estimation (MLE):
\vspace{-5pt}
\[
\LSL(\theta) 
= - \sum_{(\vecystar, \vecx) \in \YXtrain} \log p_\theta(\vecx \mid \vecystar).\vspace{-5pt}
\]
Ater SL, the model can rapidly generate many plausible designs, enabling efficient exploration in RL.

\vspace{-0.2cm}
\subsection{Reinforcement Learning (RL)}


Supervised learning aligns the model with solver-generated sequences but does not directly optimize our thermodynamic objective (e.g., Boltzmann probability $p(\vecystar \mid \vecx)$ and ensemble defect $\ED(\vecx, \vecystar)$).
We therefore fine-tune $\LM{\theta}$ with a simple group-based REINFORCE-style objective~\cite{williams:1992}, often referred to as \emph{Group Relative Policy Optimization} (GRPO)~\cite{shao+2024}.

Given a target structure $\vecystar$, we sample a group of $K$ sequences
\(
\{\vecx_k\}_{k=1}^K \sim p_\theta(\cdot \mid \vecystar)
\)
(using constrained decoding) and compute rewards $R_k = R(\vecx_k, \vecystar)$.
We form a group baseline 
\(
b = \tfrac{1}{K} \sum_{k=1}^K R_k
\)
with standard deviation
\(
\sigma = \mathrm{Std}_k(R_k),
\),
and define a normalized advantage
\(
\hat{A}_k = \frac{R_k - b}{\sigma + \varepsilon}.
\)
The GRPO loss is
\begin{equation}
\vspace{-5pt}
\mathcal{L}_{\text{GRPO}}(\theta)
= - \mathbb{E}_k \big[ \hat{A}_k \log p_\theta(\vecx_k \mid \vecystar) \big],
\end{equation}
which increases the probability of better-than-average samples within each group and decreases that of worse-than-average ones, on top of the supervised initialization.

\paragraph{Reward Design.}
For reinforcement learning, we map the thermodynamic quantities in Section~\ref{sec:prelim} to a single scalar reward that reflects how well a candidate sequence $\vecx$ folds into the target structure $\vecystar$.
During RL, we always use constrained decoding (Section~\ref{sec:decoding}) so that sampled sequences satisfy the length and base-pairing constraints of $\mathcal{X}(\vecystar)$; the reward is therefore defined over $\vecx \in \mathcal{X}(\vecystar)$.

\paragraph{Reward components.}
For each sampled pair $(\vecx, \vecystar)$, we call an RNA folding engine to obtain the Boltzmann probability $p(\vecystar \mid \vecx)$, the set of MFE structures $\MFEs(\vecx)$, and whether the MFE is unique.
Let $\mathbb{I}[\cdot]$ denote the indicator function.
We define three reward components:
\begin{equation}
\begin{small}
\begin{aligned}
r_{\text{prob}}(\vecx, \vecystar)
&= p(\vecystar \mid \vecx), \;
r_{\text{mfe}}(\vecx, \vecystar)
=\mathbb{I}\big[\vecystar \in \MFEs(\vecx)\big], \\
&r_{\text{umfe}}(\vecx, \vecystar)
= \mathbb{I}\big[\vecystar \in \text{uMFE}(\vecx)\big].
\end{aligned}\hspace{-5pt}
\end{small}
\end{equation}
The first term rewards sequences whose Boltzmann ensemble is concentrated on the target structure.
The second term adds a bonus whenever the target structure appears among the MFE structures of $\vecx$. 
The third term further rewards sequences whose MFE structure is unique, discouraging strong competing alternative folds. Here $\text{uMFE}(\vecx)$ denotes the unique MFE structure if it exists and the empty set otherwise.

\paragraph{Combined reward.}
We combine the three rewards with simple linear shaping:
\begin{equation}
\begin{aligned}
R(\vecx, \vecystar)
&= 0.5\,r_{\text{prob}}(\vecx, \vecystar)
 + 0.25\,r_{\text{mfe}}(\vecx, \vecystar) \\
&\quad + 0.25\,r_{\text{umfe}}(\vecx, \vecystar), \qquad
\vecx \in \mathcal{X}(\vecystar).
\end{aligned}
\end{equation}

By construction, $r_{\text{prob}} \in [0,1]$ and $r_{\text{mfe}}, r_{\text{umfe}} \in [0, 0.5]$, so $R(\vecx, \vecystar) \in [0,1]$ and $R(\vecx, \vecystar) = 1$ exactly when the target structure has probability $1$ and is the unique MFE structure.
If the thermodynamic evaluation fails for numerical or library-related reasons, we fall back to a neutral reward $R(\vecx, \vecystar) = 0$, so such rare cases neither strongly encourage nor discourage the policy.


\section{Datasets Construction}
\label{sec:dataset}

\subsection{Test Dataset \Ytest}

For RNA design, the test set \Ytest is a collection of target structures \vecystar.
We create a comprehensive test set consisting of four datasets, 
covering both manually-designed puzzles and native structures:

\begin{enumerate}
\item {\bf Eterna100} \cite{anderson+:2016}
is the standard and  most widely used benchmark for evaluating
RNA design programs. It contains 100 secondary structures
(i.e., human-designed ``puzzles'') of up to 400 \nts, varying in design difficulty from simple hairpins to intricate multiloop structures.


\item {\bf Eterna100-v2} \cite{koodli2021redesigning}
modifies 19 puzzles from Eterna100 so that all puzzles are MFE-solvable
under Turner 2004 energy model~\cite{mittal2024nndb} via ViennaRNA v2~\cite{lorenz2011viennarna}.

\item {\bf Rfam-Taneda-27} \cite{taneda:2011modena}
includes consensus secondary structures from seed alignments in Rfam 9.0,
with each entry corresponding to an RNA family. 

\item {\bf RNAsolo-100}, extracted from \cite{adamczyk2022rnasolo},
is a collection of secondary structures curated from a database of experimentally determined
RNA 3D structures. The specific ``100'' subset here refers to 100 cleaned, non-redundant RNA structures we selected for secondary-structure-design benchmarking.
\end{enumerate}

\vspace{-0.2cm}
We call the set of unique structures from all four datasets our {\bf ``Union Test Set''}, notated \Ytest, which contains 246 unique structures, with length up to 400 \nts.

\vspace{-0.2cm}

\subsection{Supervised Learning Dataset \YXtrain}
\label{sec:SLdata}

Our supervised training dataset \YXtrain is a set of $(\vecystar, \vecx)$ pairs where each \vecystar is a target structure (i.e., a puzzle) and each \vecx is the corresponding RNA sequence (i.e., a design).
We construct this set in two steps:
(a) collect the set of target structures \Ytrain,
and then
(b) use a state-of-the-art tool to generate candidate designs for each structure in \Ytrain.

For step (a),  while previous work 
used native or native-derived structures 
such as Rfam-Learn \cite{runge+:2019}, 
those data suffers from three serious drawbacks:
\begin{enumerate}
\item {\bf Testset contamination}: the test set contains native structures that are too similar to some native or native-derived ones in the training set;
\item {\bf Limited size}: native structures are often limited in quantity (e.g., Rfam-Learn has only 65K structures),
which may not provide enough training data for the data-hungry Transformer language models.
\item {\bf Limited diversity}: many native sequences are homologous ones from the same families (e.g., Rfam-Learn is derived from Rfam, a collection of native RNA sequences by families), where sequences and structures in each family share high sequence identities and structural similarities.
\end{enumerate}

\begin{figure}[t]
\centering
\vspace{-0.5cm}


\newcommand{\rowgap}{1.7cm}

\hspace{-0.3cm}
\resizebox{1.04\linewidth}{!}{

\begin{tikzpicture}[
    node distance=2.2cm and 1.9cm,
    font=\small,
    box/.style={
        draw,
        rounded corners,
        minimum width=0.2cm,
        minimum height=0.2cm,
        align=center
    },
    arrow/.style={->, thick, >=Stealth},
    label/.style={draw=none, align=center}
]

\node[label, font=\bfseries] (sl_title) {(a)};

\node[box, right=0.05cm of sl_title] (sl_a) {$\{\vecx\}$\\1M};
\node[label, above=0.01mm of sl_a] {Random\\Sequences};

\node[box, right=1.4cm of sl_a] (sl_b) {$\{\vecy^{\star}\}$\\1M};
\node[label, above=0.7mm of sl_b] {Random\\MFE};

\node[box, right=0.7cm of sl_b] (sl_c) {SAMFEO};

\node[box, right=0.7cm of sl_c] (sl_d) {$\{(\vecy^{\star}, \vecx)\}$\\10M};

\draw[arrow] (sl_a) -- node[midway, above]{MFE} node[midway, below]{fold} (sl_b);
\draw[arrow] (sl_b) -- (sl_c);
\draw[arrow] ([yshift=4pt]sl_c.east) -- ([yshift=10pt]sl_d.west);
\draw[arrow] (sl_c.east) -- (sl_d.west) node[midway, below, yshift=-8pt]{$\times 10$};
\draw[arrow] ([yshift=-4pt]sl_c.east) -- ([yshift=-10pt]sl_d.west);

\node[label, below=0.05mm of sl_b] {$\Ytrain$};
\node[label, below=0.05mm of sl_d] {$\YXtrain$};

\node[label, font=\bfseries, below=\rowgap of sl_title] (rl_title) {(b)};

\node[box, right=0.05cm of rl_title] (rl_a) {$\{ \vecy^{\star} \}$\\22K};
\node[label, above=0.7mm of rl_a] {Eterna-web};

\node[box, right=1.4cm of rl_a] (rl_b) {$\{ \vecy^{\star} \}$\\18.3K};
\node[label, above=0.01mm of rl_b] {Eterna-web\\designable};
\node[label, below=0.05mm of rl_b] {$\YRLraw$};

\node[box, right=2.cm of rl_b] (rl_c) {$\{ \vecy^{\star} \}$\\13.1K};
\node[label, below=0.05mm of rl_c] {$\YRLlarge$};

\node[box, right=2.1cm of rl_c] (rl_d) {$\{\vecy^{\star}\}$\\2.8K};
\node[label, below=0.05mm of rl_d] {$\YRL$};

\draw[arrow] (rl_a) -- node[midway, above]{designable} (rl_b);
\draw[arrow] (rl_b) -- node[midway, above]{$d_{\text{min\_norm}}\! > \! 0.2$} (rl_c);
\draw[arrow] (rl_c) -- node[midway, above]{AoN$(\vecy^{\star}) \geq 0.1$} node[midway, below, align=center]{NSD$(\vecy^{\star}) > 0.5$ } (rl_d);

\end{tikzpicture}
}
\vspace{-0.5cm}
\caption{Workflows for constructing (a) SL and (b) RL training datasets. Here, $d_{\text{min\_norm}}$ denotes $d_{\text{min\_norm}}(\vecy^{\star}, \Ytest)$.}
\label{fig:workflow}
\smallskip
\end{figure}
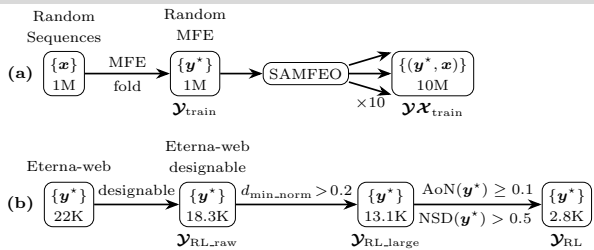

To addresses these issues,
we instead propose to use large amounts of randomly generated (and thus diverse) artificial structures
unrelated to the test sets.
However, directly generating random secondary structures 
is problematic, as 
most of them will be undesignable in the MFE criteria \cite{zhou+:2024}.
Instead, we first generated 1M random RNA {\em sequences} of lengths up to 500 \nts, with uniform distribution of nucleotides $\{\nucA,\nucC,\nucG,\nucU\}$, and then use ViennaRNA v2 to get their MFE structures; now we have 1M random-MFE structures in \Ytrain  (see Fig.~\ref{fig:workflow}(a)). 
This way we made sure all these structures are designable by the MFE criterion.\footnote{Although some of \Ytrain might not be uMFE-designable.}

For each target structure in \Ytrain, 
we use the state-of-the-art SAMFEO tool 
to generate 10 candidate designs,
so finally we have 10M $(\vecystar,\vecx)$ pairs in \YXtrain.
Surprisingly, this random-induced supervised dataset unrelated to the test set yields competitive results on the latter (Sec.~\ref{sec:sl_results}).

%
%

\subsection{Reinforcement Learning Dataset \YRL}
\label{subsec:yrl_selection}

It is a general consensus in language model training
that the later the training stage,
the less data is needed (with more compute per example), 
but with higher quality standards.
Given that RL is the last stage in training,
we construct a small and relevant dataset \YRL in three steps.
It is important to note that, different from supervised learning, RL only needs  structures \vecystar,
but not solutions \vecx,
since it will explore possible solutions via decoding.

\paragraph{Step 1.  data source: Eterna-web.}
We first downloaded the full collection of 22K player-designed structures (i.e., puzzles) from the Eterna website (\url{http://eternagame.org}), excluding those from Eterna100. 
We then removed structures longer than 500 \nts 
(to match test sets), and removed MFE-undesignable ones using
our previous work \cite{zhou+:2024},
which resulted in 18.3K MFE-designable structures (\YRLraw in Fig.~\ref{fig:workflow}(b)).

\begin{figure*}[t]
  \hspace{-0.2cm}
  \setlength{\tabcolsep}{-1pt}
  \begin{tabular}{cccc}
    \includegraphics[width=0.24\linewidth]{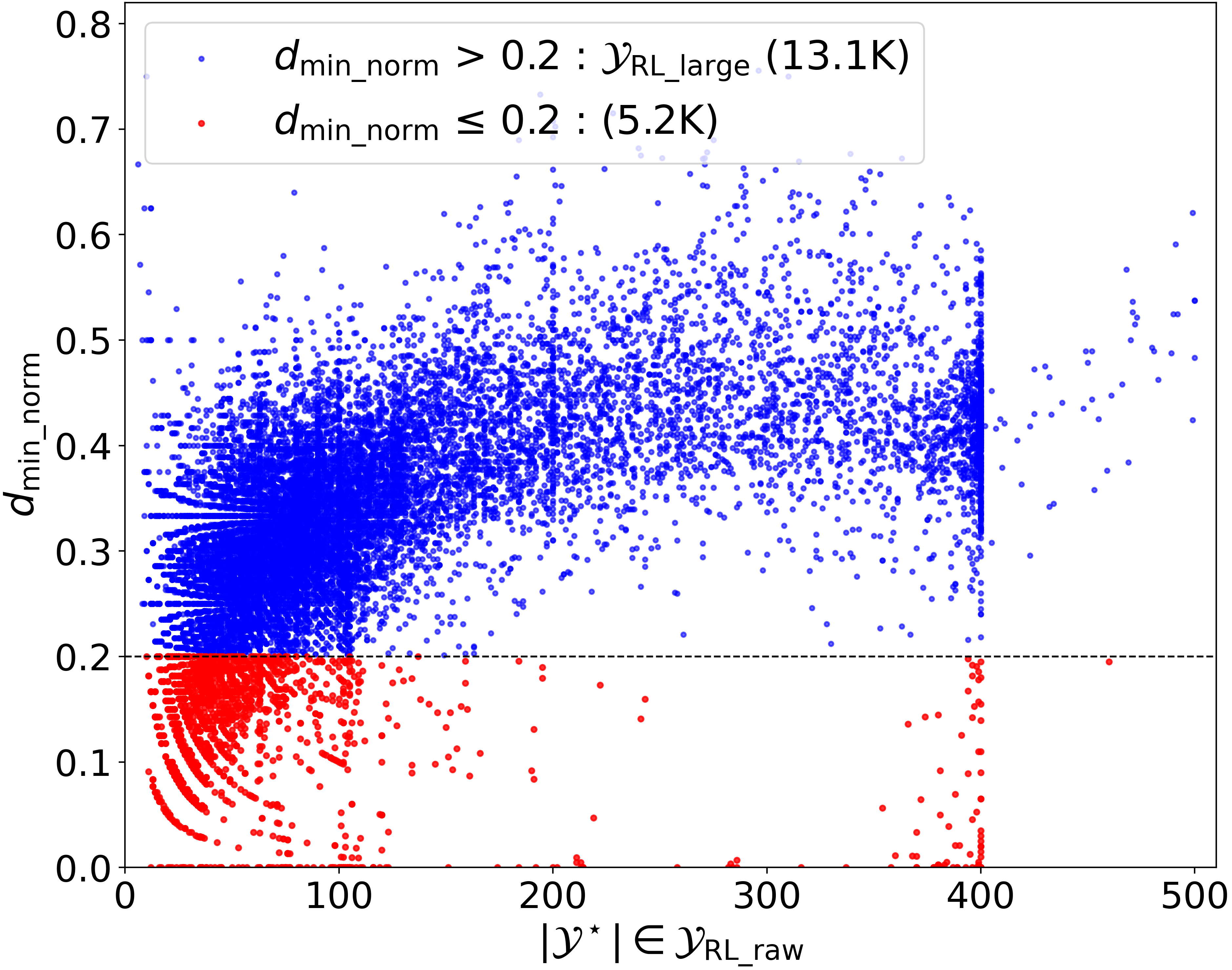}
     &
    \hspace{-0.5cm}\includegraphics[width=0.255\linewidth]{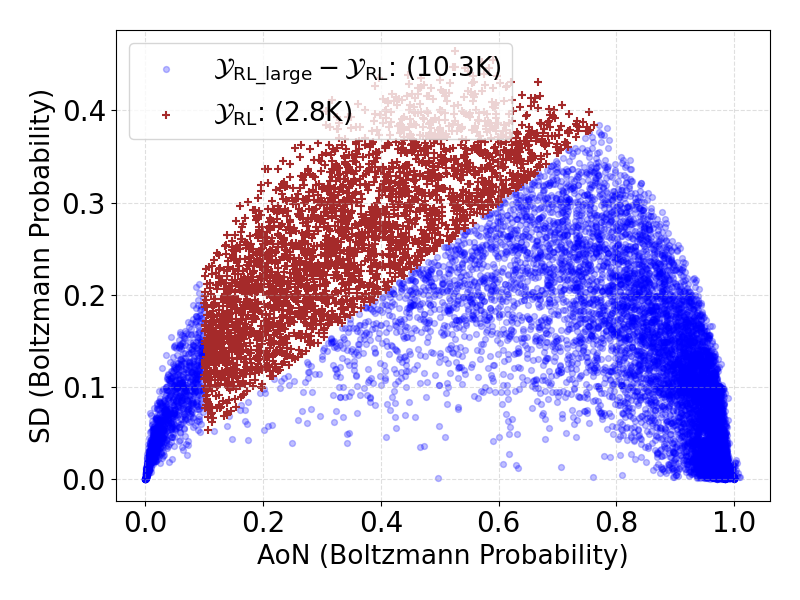}\hspace{-0.5cm}
    &
    \includegraphics[width=0.255\linewidth]{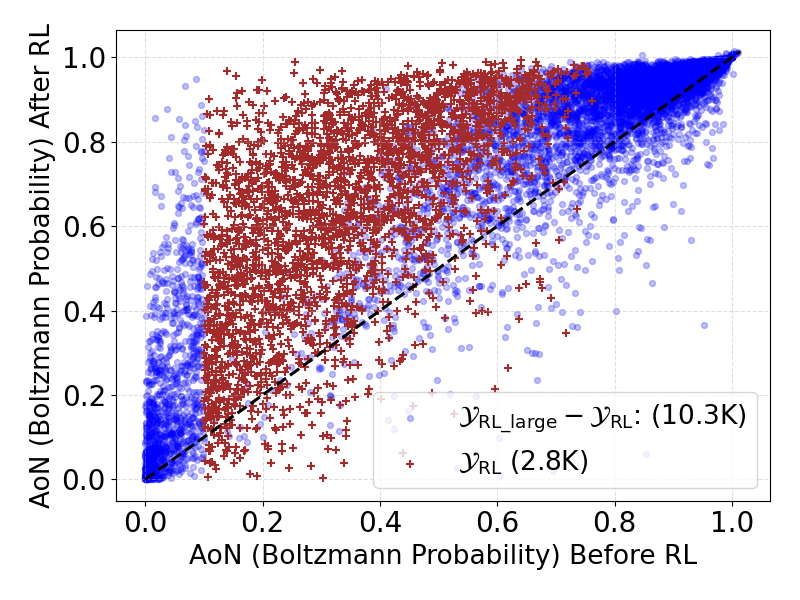}\hspace{0.2cm}
    &
    \includegraphics[width=0.255\linewidth]{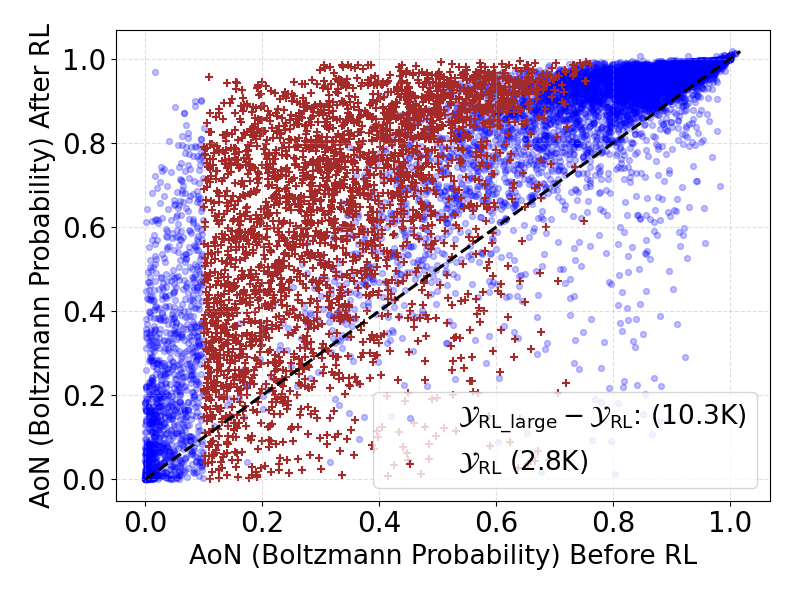}
    \\
    {\tiny (a) distance filtering on \YRLraw} & \hspace{0.01cm} {\tiny (b) decoding \YRLlarge with SL model}   & {\tiny (c) RL on \YRL}   & {\tiny (d) RL on \YRLlarge}
  \end{tabular}
\caption{RL dataset selection. (a) Removing structures in \YRLraw that are too close to the Eterna100 testset. (b) Selecting the small subset \YRL from \YRLlarge. 
(c--d) Decoding results on \YRLlarge using RL models trained on \YRL and \YRLlarge.  \label{fig:rl_data}}
\end{figure*}

\paragraph{Step 2. Removing structures similar to test data.}
Unlike our SL training set \YXtrain
of random-induced structures that are unrelated to our test data,
our RL set is drawn from a distribution similar to one of our testsets (Eterna100), 
so potentially there is a risk of testset contamination.
To ensure a rigorous evaluation, we removed structures 
$\vecystar \in \YRLraw$ that were too close to any test structure 
$\vecy'$ in $\Ytest$. For each \vecystar, we computed a 
\textit{normalized minimum edit distance} that accounts for length differences:
\[
\vspace{-0.15cm}
d_{\text{min\_norm}}(\vecystar, \Ytest) =
\min_{\vecy' \in \Ytest}
\frac{d_{\text{edit}}(\vecystar, \vecy')}{\min(|\vecystar|, |\vecy'|)}.
\]
Here, $d_{\text{edit}}(\cdot, \cdot)$ denotes the edit distance between two structures, 
and normalization by the length of the shorter structure 
$\min(|\vecystar|, |\vecy'|)$ ensures that our similarity measure 
is comparable across structures of different lengths. 
Structures in \YRLraw with a $d_{\text{min\_norm}}(\vecystar, \Ytest)$ 0.2 or less (i.e., 80\%+ similar) were excluded to 
prevent information leakage between RL and test sets (Fig.~\ref{fig:rl_data}(a)),
resulting in 13.1K structures (\YRLlarge in Fig.~\ref{fig:workflow}(b)).

\paragraph{Step 3. Further filtering for RL efficiency and quality.}
We could do RL on the above \YRLlarge set, 
but it is computationally expensive because, 
for each target structure 
the model must generate many samples (``exploration'') and evaluate their folding performance. 
Our goal is to identify a small subset \(\YRL \subseteq \YRLlarge\) for which RL is likely to yield meaningful improvement.


To do so, we first decode structures in \YRLlarge with the SL model trained on
 \YXtrain. For each structure $\vecystar$ in $\YRLlarge$, we generate $K$ candidate sequences
$\{\vecx_k\}_{k=1}^K$ and compute their Boltzmann probabilities
$p(\vecystar \mid \vecx_k)$. We then summarize the distribution of these probabilities
with two statistics:

\begin{itemize}
  \item \textbf{Average probability (AoN):} \(\mathrm{AoN}(\vecystar) = \frac{1}{K}\sum_{k=1}^{K} p(\vecystar \mid \vecx_k)\), which reflects the performance of the SL model for structure in terms of the ensemble metric $p(\vecystar \mid \vecx_k)$.
\vspace{2mm}
  \item \textbf{Standard deviation of probabilities:}
  \(\mathrm{SD}(\vecystar) = \mathrm{Std}_k\!\left(p(\vecystar \mid \vecx_k)\right)\) capturing the variability, and hence the potential for improvement across sampled designs.
\end{itemize}

Using $\mathrm{SD}(\vecystar)$ alone can be misleading for structures whose probabilities remain uniformly low.
Thus, we propose the \emph{normalized standard deviation} (NSD):
 \( \mathrm{NSD}(\vecystar) = \frac{\mathrm{SD}(\vecystar)}{\mathrm{AoN}(\vecystar)},\)
which expresses variability relative to the mean probability and highlights structures where RL can exploit
useful diversity in folding outcomes.

\paragraph{Selection criteria.}
We select a structure $\vecystar$ from $\Ytrain$ for reinforcement learning if

\vspace{-8pt}
\begin{equation}
  \mathrm{AoN}(\vecystar) \ge 0.1
  \quad\text{and}\quad
  \mathrm{NSD}(\vecystar) > 0.5.
\end{equation}
The AoN threshold removes structures on which the SL model performs extremely poorly, while the NSD threshold ensures that there is sufficient variance among sampled sequences for RL to meaningfully refine the model. Applying both criteria yields a filtered subset \YRL of approximately 2.8K structures (Fig.~\ref{fig:rl_data}(b)) on which RL is both computationally feasible and empirically effective. Moreover, Fig.~\ref{fig:rl_data}(c--d) compare AoN scores before and after RL when training on either the full set \YRLlarge or the filtered subset \YRL. When RL is applied to the entire \YRLlarge, many structures in $\YRLlarge$ actually exhibit degraded performance after RL. In contrast, when RL is restricted to \YRL, the model not only improves performance on most structures in $\YRL$ but also generalizes positively, yielding improvements for many structures in \YRLlarge as well.
These results confirm that our filtering yields a small subset 
that improves both the efficiency and quality of RL training.
Sec.~\ref{sec:efficiency} further showed a $\sim2.9\times$ speedup by RL on \YRL (16 hours vs.~47 hours of RL on \YRLlarge).

\section{Evaluation Results}
\label{sec:eval}

\subsection{Supervised Learning Results}
\label{sec:sl_results}

\begin{figure}[t]
    \centering
    \includegraphics[width=.9\linewidth]{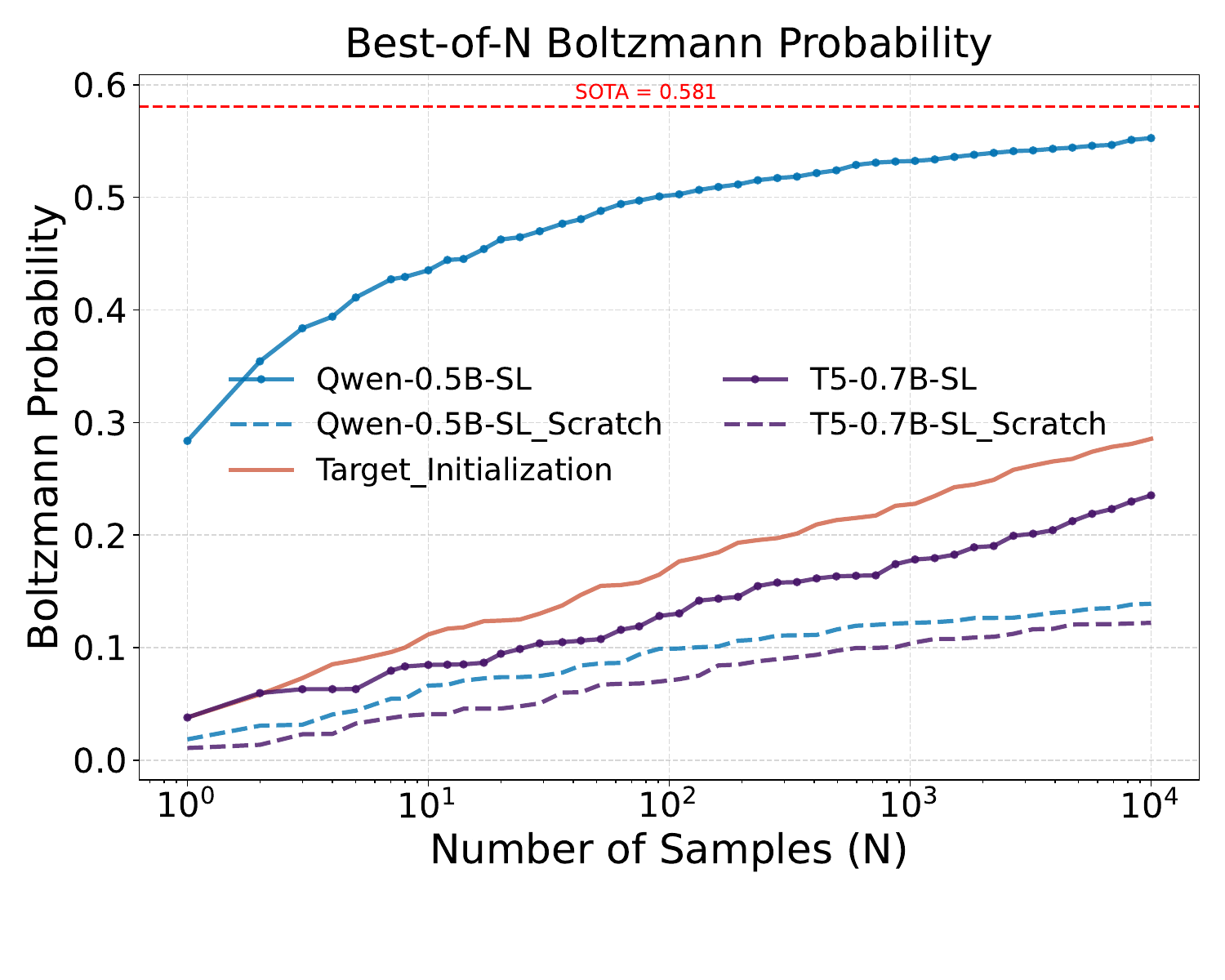}
\vspace{-0.8cm}
    \caption{Supervised learning results of different models trained on \YXtrain in terms of best-of-$N$ Boltzmann probability on Eterna100. 
    \label{fig:sl_plot_with_table}}
\end{figure}

Figure~\ref{fig:sl_plot_with_table}(a) reports the best-of-$N$ Boltzmann probability achieved by each supervised-learning (SL) model trained on our 10M \YXtrain set as a function of the number of decoded samples $N$ per target structure. Across all sampling budgets, the pretrained Qwen-0.5B-SL model performs the strongest after SL, reaching a best-of-$N$ probability of approximately $0.55$ at $N=10^4$. 
By contrast, both models trained from scratch, Qwen-0.5B-SL\_Scratch (decoder-only) and T5-0.7B-SL\_Scratch (encoder-decoder), perform substantially worse. 
These results clearly suggest that training Transformer LMs for RNA design entirely from scratch is ineffective regardless of architecture. The \texttt{Target\_Initialization} 
is a simple baseline where for each unpaired position in \vecystar,
we use a distribution of 90\% \nucA\/ (other nucleotides 
for the remaining 10\%)
and for each basepair in \vecystar,
we use a distribution of 70\% \nucG\nucC/\nucC\nucG\/ pairs, 40\% \nucA\nucU/\nucU\nucA\/ pairs, and 10\% Wobble pairs.
This baseline performs surprisingly well and even surpasses both T5 models at moderate sampling budgets. 
Although it 
ultimately is far below the best model, its competitiveness underscores the limitations of uninitialized models in capturing structural constraints.

On the other hand, it is surprising and encouraging that supervised training on random-induced structure-sequence pairs \YXtrain which are completely unrelated to the test set can perform reasonably well on the latter.



\vspace{-0.4cm}

\subsection{Decoding: Constrained vs.~Unconstrained}
We empirically compare naive decoding and constrained decoding on the union test set \Ytest using the 
SL-trained Qwen2.5-0.5B-SL model.  
For each target structure, we draw $10^3$ samples with both decoding methods and for every decoded sequence, record 
(i) structural invalidity (any base pair in $\vecystar$ violating Watson-Crick-Franklin or wobble rules), 
and (ii) wall-clock decoding time.

\begin{figure}[h] 
    \centering
    \includegraphics[width=0.975\linewidth]{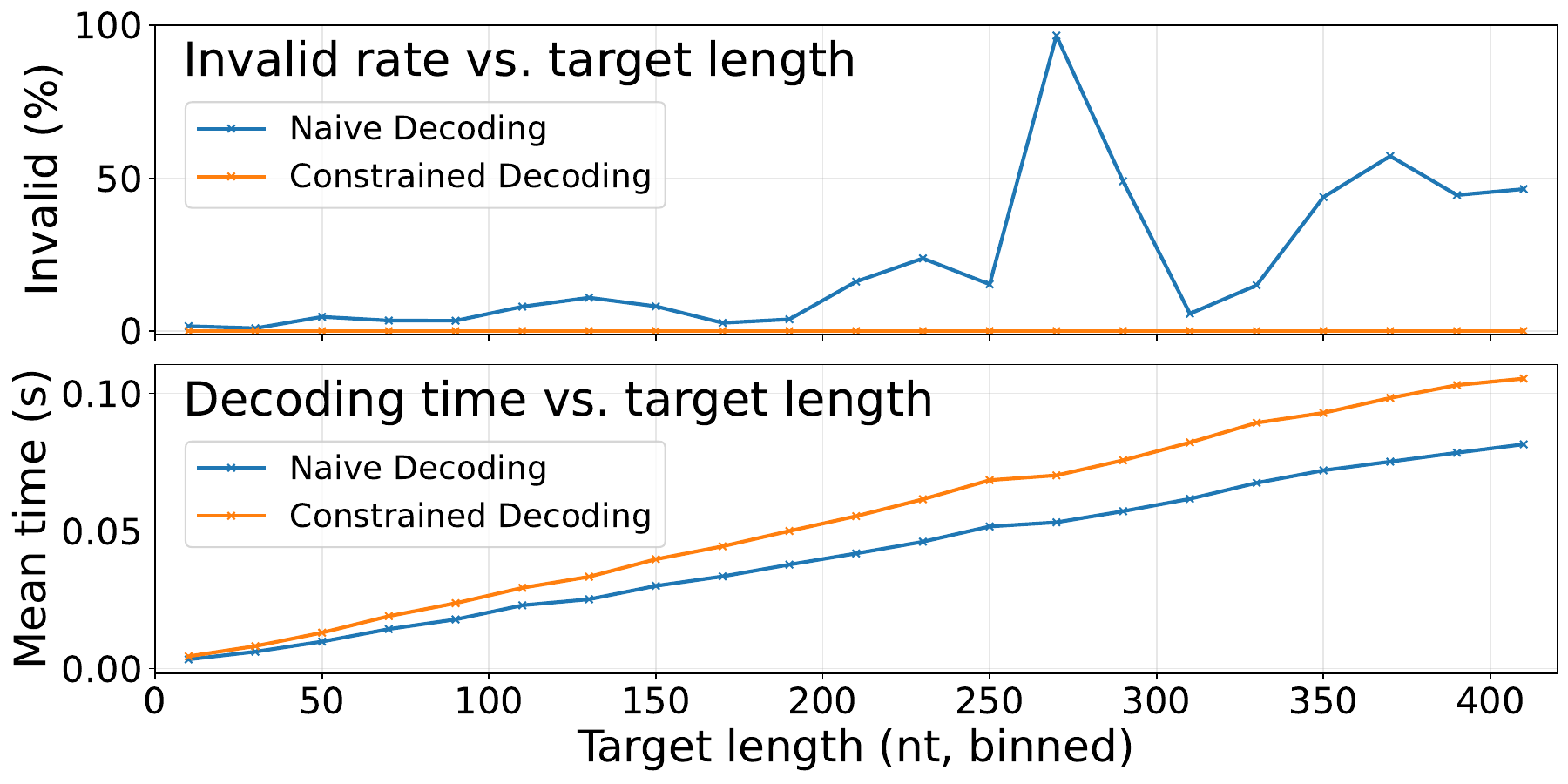}
    \vspace{-8pt}
    \caption{Naive decoding vs.~constrained decoding on the Union Test Set \Ytest($10^3$ samples per structure; supervised learning model). Naive decoding becomes increasingly invalid for longer targets, while constrained decoding guarantees validity at a $\sim30\%$ slowdown.}
    \label{fig:combined_metrics_vs_length}
\end{figure}

As shown in Fig.~\ref{fig:combined_metrics_vs_length}, naive decoding yields a sharp rise in invalid sequences as target length increases, wasting a substantial fraction of samples.  
In contrast, constrained decoding eliminates invalid outputs by design,
while incurring a modest $\sim30\%$ 
overhead. 
This modest overhead is more than offset by the improved effective sample efficiency, so we adopt constrained decoding in RL.

Fig.~\ref{fig:attention-map} additionally shows an attention map from constrained decoding, which
correlates with base-pairs \vecystar. 

\vspace{-15pt}

\smallskip
\subsection{Reinforcement Learning Results}
\label{sec:rl_results}

\begin{figure*}[h]
  \centering
  \vspace{-0.4cm}
  \hspace{-0.3cm}
  \begin{tabular}{ccc}
    \includegraphics[height=4.85cm, width=5cm]{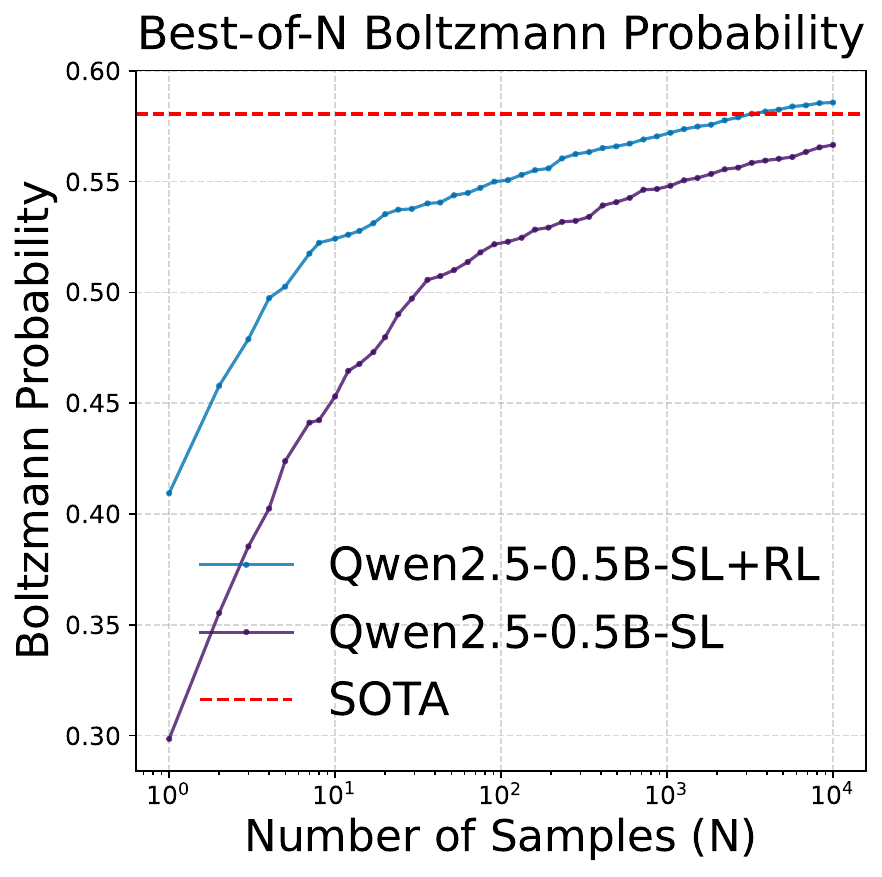} 
    &
    \includegraphics[height=4.85cm, width=5cm]{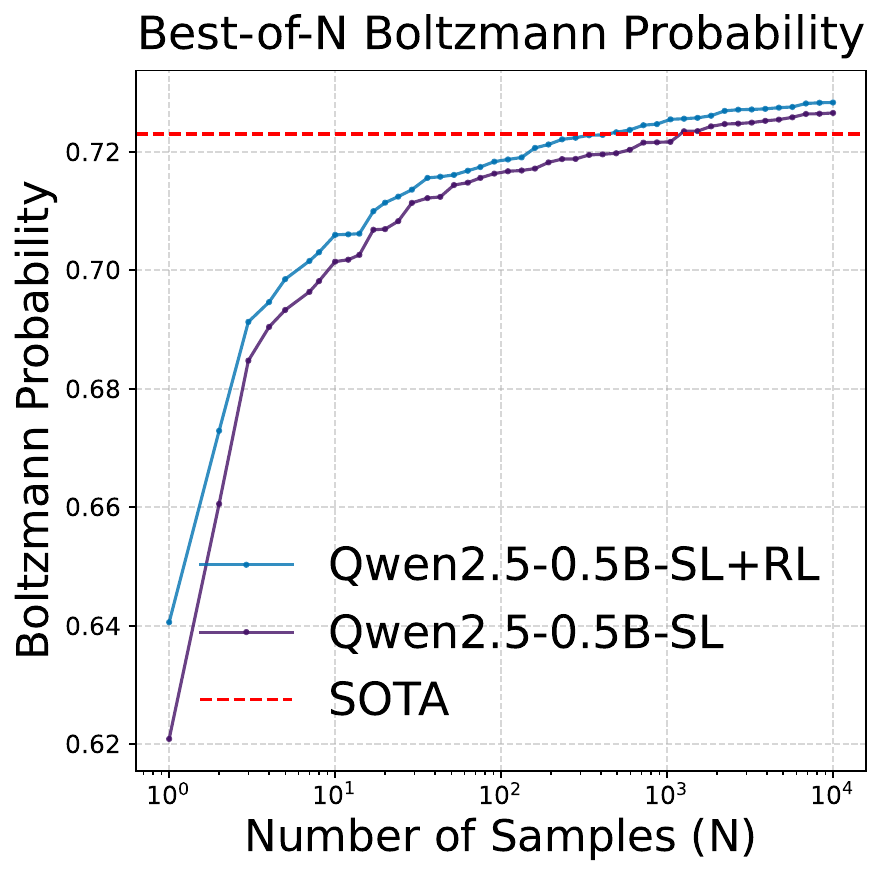} 
    &
    \includegraphics[height=4.85cm, width=5cm]{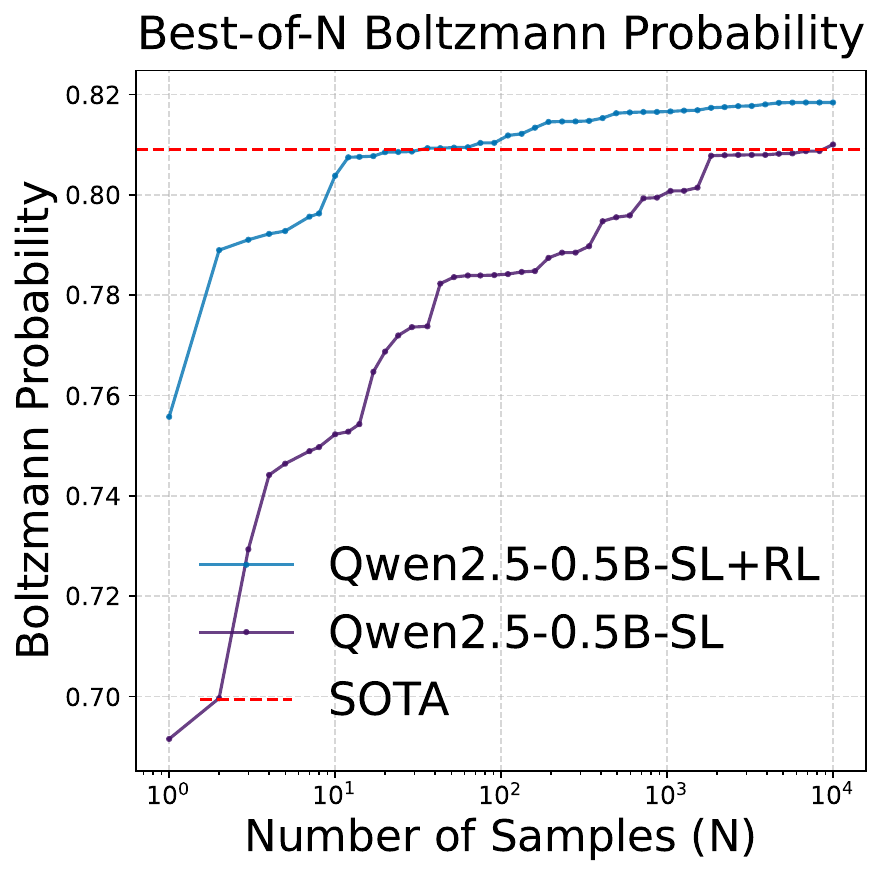}
    \\[-0.1cm]
    (a) Eterna100 & (b) RNAsolo100 & (c) Rfam-Taneda-27
  \end{tabular}
  \vspace{-0.2cm}
\caption{Best-of-$N$ probability curves for the SL and SL+RL models on three test sets.
In all cases,
RL brought substantial improvements over SL,
surpassing SOTA (SAMFEO); in particular, RL only needs $\sim$15 samples to surpass SAMFEO in (c).
See Fig.~\ref{fig:SI-SL+RL} for other metrics.\label{fig:SL+RL}}
\end{figure*}

Fig.~\ref{fig:SL+RL} shows that applying reinforcement learning (RL) on top of the 
supervised-learning (SL) model leads to a substantial improvement in best-of-$N$ Boltzmann probability 
across all three benchmarks. The RL-enhanced model consistently outperforms the SL-only model, with 
particularly large gains on Eterna100 and Rfam-Taneda-27, where the performance gap remains pronounced 
for all sampling budgets. 
RL greatly improves sample efficiency especially on Rfam-Taneda-27, where only 15 samples are needed to surpass SAMFEO in Boltzmann probability.

Overall, these results demonstrate that RL significantly strengthens the model’s ability to generate 
high-probability designs, especially for benchmarks involving complex or human-designed structures, and 
consistently improves the scaling behavior of the SL baseline. Similarly, from Table~\ref{tab:comparison-eterna100-rnasolo100}, we observe that our reinforcement-fine-tuned model consistently matches or surpasses other methods on ensemble-based metrics. It also achieves the best performance on $p(\vecystar \mid \vecx)$ accross all datasets and attains the best NED—tied only with SAMFEO on RNAsolo100, while outperforming RNAinverse-pf (updated in 2023) and NEMO on all datasets for both $p(\vecystar \mid \vecx)$ and NED. Similarly, our SL+RL model also demonstrates high sample efficiency in generating MFE solutions. As shown in Fig.~\ref{fig:SI-SL+RL_Solve_rate}, our model requires only 2 samples on Rfam-Taneda-27 and 50 samples on RNAsolo100 to match the performance of the state-of-the-art method.

\begin{table}
\centering
\hspace{-5pt}
\resizebox{.486\textwidth}{!}{
\setlength{\tabcolsep}{1pt}
\begin{tabular}{l|cccc|cccc}
\hline
\multirow{2}{*}{Method} 
    & \multicolumn{4}{c|}{\textbf{Eterna100}}
    & \multicolumn{4}{c}{\textbf{RNAsolo100}} \\
\cline{2-9}
    & \small $p(\vec{y}^\star|\vec{x})$ 
    & \small NED$_\downarrow$
    & \small MFE 
    & \small uMFE 
    & \small $p(\vec{y}^\star|\vec{x})$ 
    & \small NED$_\downarrow$
    & \small MFE 
    & \small uMFE\\ 
\hline
RNAinverse-pf 
    & 0.571 & 0.045 & 77 & 72
    & 0.710 & 0.015  & 79 & 79 \\
NEMO 
    & 0.271 & 0.098 & \textbf{79} & \textbf{77}
    & 0.506 & 0.038 & \textbf{81} & \textbf{81} \\
SAMFEO 
    & 0.580 & 0.043 & 77 & 74
    & 0.723 & \textbf{0.013} & 80 & 79 \\
\hline
{Ours: SL+RL} 
    & \textbf{0.586} & \textbf{0.042} & 74 & 72
    & \textbf{0.728} & \textbf{0.013} & \textbf{81} & \textbf{81} \\
\hline
\end{tabular}
}
\centering
\resizebox{.486\textwidth}{!}{
\setlength{\tabcolsep}{1pt}
\hspace{-5pt}
\begin{tabular}{l|cccc|cccc}
\hline
\multirow{2}{*}{Method} 
    & \multicolumn{4}{c|}{\textbf{Eterna100-v2}} 
    & \multicolumn{4}{c}{\textbf{Rfam-Taneda-27}} \\
\cline{2-9}
    & \small $p(\vec{y}^\star|\vec{x})$ 
    & \small NED$_\downarrow$
    & \small MFE 
    & \small uMFE 
    & \small $p(\vec{y}^\star|\vec{x})$ 
    & \small NED$_\downarrow$
    & \small MFE 
    & \small uMFE\\ 
\hline
RNAinverse-pf 
    & 0.590 & 0.039 & 82 & 77
    & 0.808   & 0.004   & 24 & 24 \\
NEMO 
    & 0.270 & 0.098 & \textbf{90} & \textbf{87}
    & 0.411 & 0.025 & \textbf{25} & \textbf{25} \\
SAMFEO 
    & 0.596 & \textbf{0.036} & 83 & 79
    & 0.809 & 0.004 & 24 & 24 \\
\hline
{Ours: SL+RL} 
    & \textbf{0.606} & \textbf{0.036} & 81 & 79
    & \textbf{0.818} & \textbf{0.039} & 24 & 24 \\
\hline
\end{tabular}
}
\vspace{5pt}
\caption{Comparison with state-of-the-art systems on
all four test sets. RNAinverse-pf was updated in ViennaRNA 2.6 (2023). Our results report the best of \(
10^4
\)
 samples.
\label{tab:comparison-eterna100-rnasolo100}}
\end{table}

Fig.~\ref{fig:x-y plot}  presents an comparison of Boltzmann probabilities between SAMFEO (Sota) and Ours (SL+RL) model on the Eterna100 test set. Each point corresponds to a puzzle, and the size of the ``+'' marker reflects the structure length, with larger markers indicating longer puzzles. Our approach consistently achieves higher probabilities, particularly for longer and hard-to-design puzzles. The zoomed-in region highlights the most challenging puzzles, where our model shows the most substantial advantage over SAMFEO.

\subsection{Efficiency}
\label{sec:efficiency}
On a single NVIDIA H100 GPU,
our model based on Qwen2.5-0.5B requires 50 hours of supervised learning (SL) on the 10M structure-sequence pairs (\YXtrain), and 16 and 47 hours of reinforcement learning (RL) on \YRL\ and \YRLlarge 
(2.8K and 13.1K structures), respectively.

\begin{figure}[t]
    \centering
    \includegraphics[width=\linewidth]{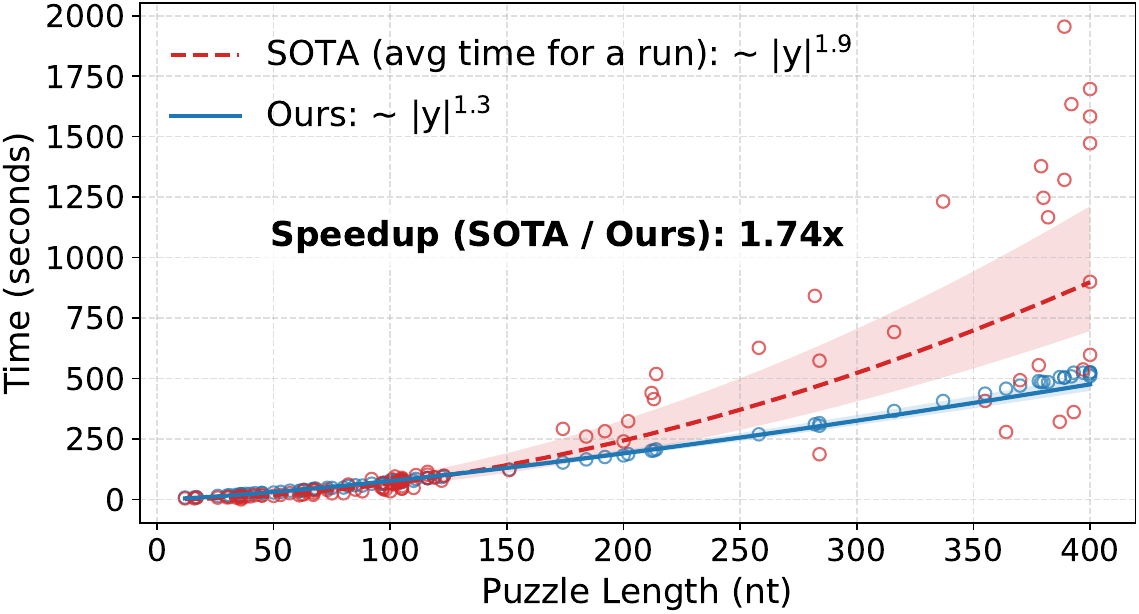}
\vspace{-0.8cm}
    \caption{Efficiency comparison with the SOTA tool SAMFEO. Most of our runtime is on GPUs (only 2\% is on CPU). Ours is 1.7$\times$ faster.}
    \label{fig:sota-vs-ours-runtime}
\end{figure}

While training might be slow, decoding is fast even with large sample size ($10^4$).
Fig.~\ref{fig:sota-vs-ours-runtime} reports decoding time on  Eterna100 as a function of structure length. 
Our method (98\% compute on GPUs and 2\% on CPU to evaluate the decoded sequence) is consistently faster than the SOTA tool SAMFEO
(with 5 parallelized runs on CPUs), and has a slower empirical complexity
(roughly $|y|^{1.3}$ vs.~$|y|^{1.9}$).  Overall, our approach yields a $1.74\times$ speedup.

\vspace{-0.2cm}
\section{Related Work}

\label{sec:related}
\paragraph{RNA LMs for RNA Analysis.} 
Several recent studies 
train BERT-style encoders on large RNA corpora for downstream 
structure-prediction and analysis,
including \textsc{RNABERT, RNA-FM, RNA-MSM}, and RiNALMo
%
\cite{RNABERT,RNA-FM,RNA-MSM,RiNALMo}.  
Benchmarks~\cite{zablocki+:2025} show that such RNA LMs improve analysis, but they (a) operate on RNA sequences without structure tokens, and (b) are encoder-only architectures  and thus incapable of {\em de novo} design.
By contrast, we use a decoder-only LM that conditions on dot–bracket structures and is trained end-to-end to \emph{generate} RNA sequences, with constrained decoding and RL explicitly targeting folding success.

\paragraph{Generative Models for RNA Generation.}
Deep generative models have recently been explored for RNA sequence design.  
GAN- and VAE-style approaches such as RNAGEN and RfamGen generate family- or target-specific RNAs 
\cite{RNAGEN,RfamGen}.  
Closer to our setting, GenerRNA is an autoregressive LM pretrained on large-scale RNA data and fine-tuned to generate functional (e.g., protein-binding) RNAs
but without conditioning on an explicit input~\cite{GenerRNA}.
RNAtranslator models protein-conditional RNA design as a sequence-to-sequence problem, where an encoder processes a protein sequence and a decoder generates RNA binders 
~\cite{RNAtranslator}.  
%
Our LM differs from these generative approaches in three ways.  
First, we condition directly on 
secondary structures.
Second, we enforce base-pairing constraints 
via constrained decoding 
rather than filtering post hoc.
Third, we fine-tune the LM with RL on 
rewards 
that align with  
standard RNA design objectives. 

\paragraph{Reinforcement Learning for RNA Design}.
RL provides an appealing framework for RNA design by framing it as a sequential decision process: at each step, an agent chooses an action (an edit on the sequence) and receives a reward based on how well the new sequence folds to the target structure.  
\namecite{eastman+:2018} pioneered this idea,
with each state being a full sequence and each action being an edit on one or two nucleotides.
\namecite{runge+:2019} introduced LEARNA and Meta-LEARNA, 
where each state is a partial sequence 
and each action is to add one unpaired or two paired nucleotides. 
\namecite{obonyo+:2022} proposed RNASP, a self-play RL framework inspired by AlphaZero. 
Partial RNA design~\cite{runge+:2024} extends LEARNA to 
handle sequence constraints.

Our use of RL is complementary and operates in a different regime.  
Rather than training a task-specific RL agent from scratch, we fine-tune a \emph{single} structure-conditioned LM that has already been trained by supervised learning on solver-generated examples.  
As a result, our model acts as an amortized neural solver: once trained, it can rapidly generate high-quality designs for new targets without per-instance RL optimization, while still benefiting from RL’s ability to directly optimize folding-based rewards (see also Tab.~\ref{tab:SI-RL}).

\vspace{-0.5cm}
\section{Conclusions and Future Work}
\label{sec:conclusions}

\vspace{-0.2cm}
We have presented the first study of using autoregressive language models for RNA design (inverse folding).
Combining pretraining, constrained decoding, supervised learning, and reinforcement learning, our final model outperforms (or ties with) 
the state-of-the-art 
in ensemble-based metrics across all four datasets, while being $1.7\times$ faster.
We also constructed and released carefully curated datasets for 
SL and RL training and testing, 
which will benefit future studies.

Our work can be improved in the following ways:
\vspace{-0.2cm}
\begin{itemize}
\item Our  decoding algorithm is sampling with context-free constraints. Since we need to sample a large amount of designs both for reinforcement learning and for testing, decoding speed 
need improvement. First, we can use other decoding algorithms such as beam search \cite{huang+:2017}.
Secondly, the $\sim$30\% overhead due to constraints could be reduced by advanced techniques for LM generation of context-free languages \cite{dong+:2024}.
\item Currently,  reinforcement learning is 
still slow, limiting us to a small sample size. 
Once decoding speed improves, we can afford to 
do better reinforcement learning with more data and 
more exploration.
\item Finally, we can explore Test-Time Reinforcement Learning (TTRL) methods
\cite{zuo+:2025} to train a different model for each testing structure, further exploiting test-time scaling.
\end{itemize}

\vspace{-0.8cm}
\bibliographystyle{natbib}
\bibliography{references}

\clearpage

\pagenumbering{roman}


\begin{appendices}

\onecolumn
\begin{figure*}
\centering
{\Large\bf Appendix: Designing RNAs with Language Models}\\[0.2cm]
M.~Gautam, N.~Dai, T.~Zhou, B.~Xie, D.~Mathews, and L.~Huang\\
Oregon State University and University of Rochester\\[0.2cm]
\end{figure*}



\setcounter{figure}{0}
\renewcommand{\thefigure}{S\arabic{figure}} 

\setcounter{table}{0}
\renewcommand{\thetable}{S\arabic{table}} 

\setcounter{section}{0}
\renewcommand{\thesection}{S\arabic{section}} 
\renewcommand{\thesubsection}{S\arabic{section}.\Alph{subsection}} 

\setcounter{footnote}{0}

%

\begin{figure*}
\centering
  \hspace{-0.2cm}
  \begin{tabular}{ccc}
    \includegraphics[width=0.3\linewidth]{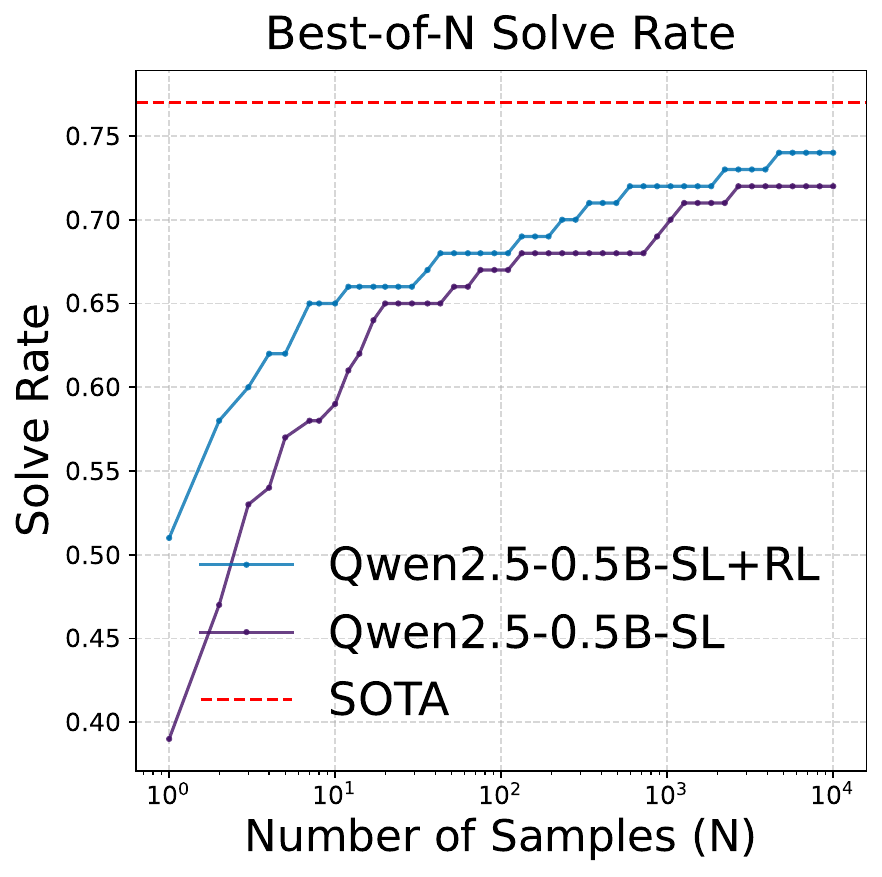}
    &
    \includegraphics[width=0.3\linewidth]{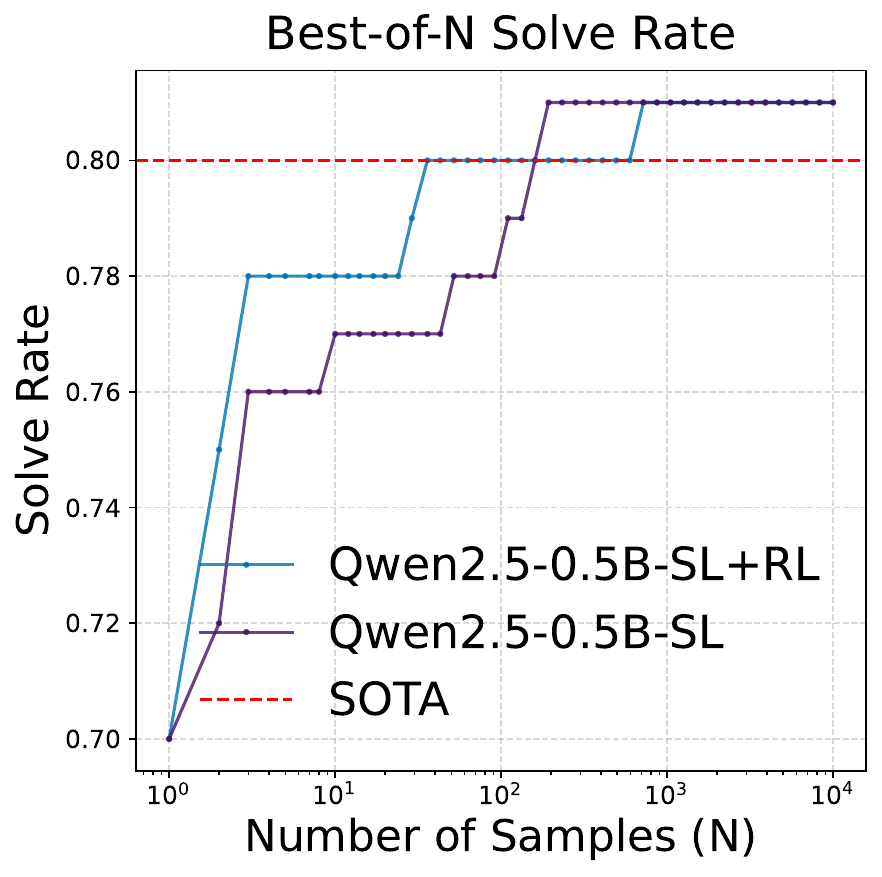}
    &
    \includegraphics[width=0.3\linewidth]{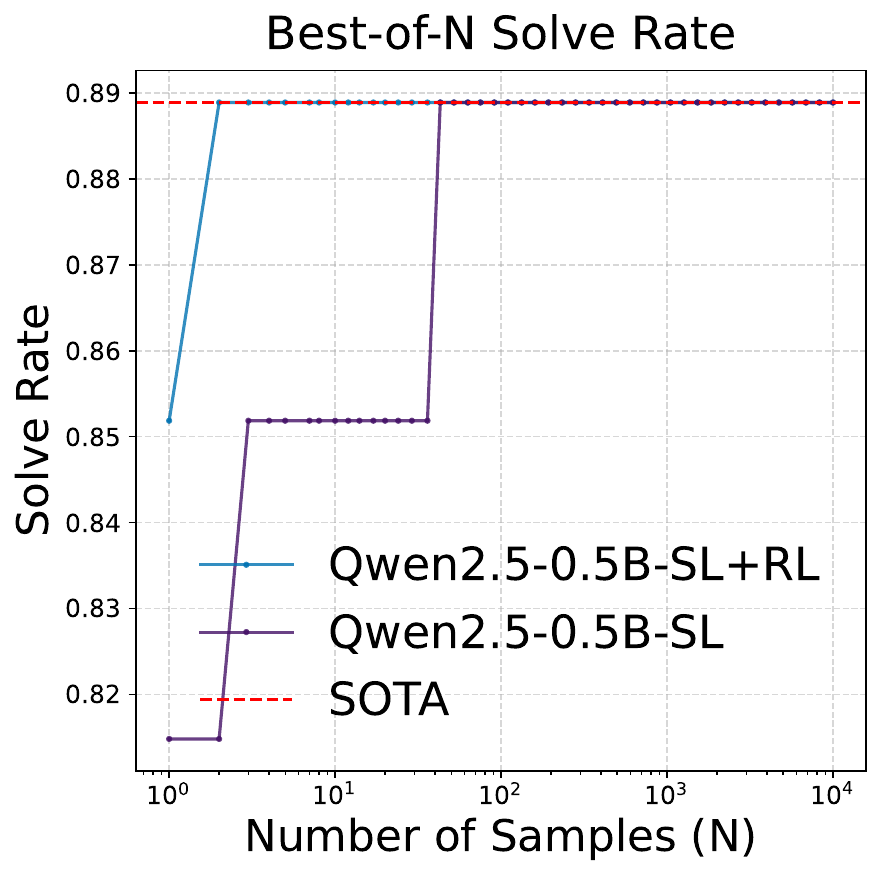}
    \\[-0.2cm]
    (a) Eterna100 & (b) RNAsolo100 & (c) Rfam-Taneda-27
  \end{tabular}
\caption{Best-of-$N$ curves for SL and SL+RL models in terms of MFE solve rate on three data sets. See Fig.~\ref{fig:SL+RL} in the main text for the corresponding Best-of-$N$ curves on Boltzmann probability.}
\label{fig:SI-SL+RL_Solve_rate}
\end{figure*}

\begin{figure*}
\centering
  \hspace{-0.2cm}
  \begin{tabular}{ccc}
    \includegraphics[width=0.3\linewidth]{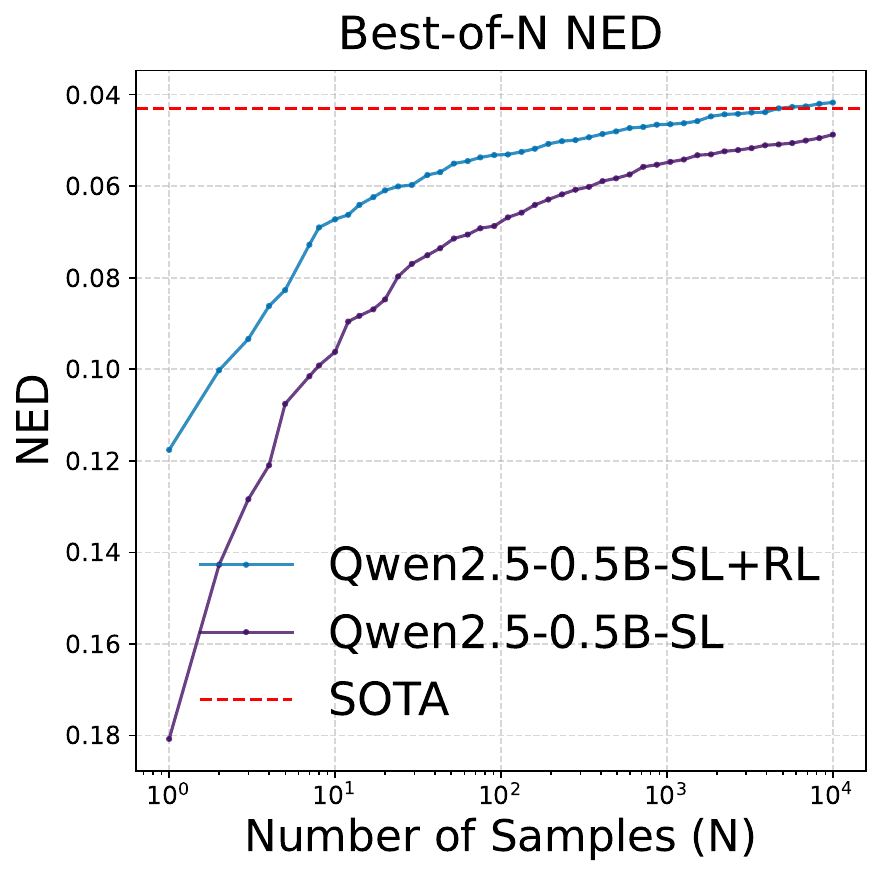}
    &
    \includegraphics[width=0.3\linewidth]{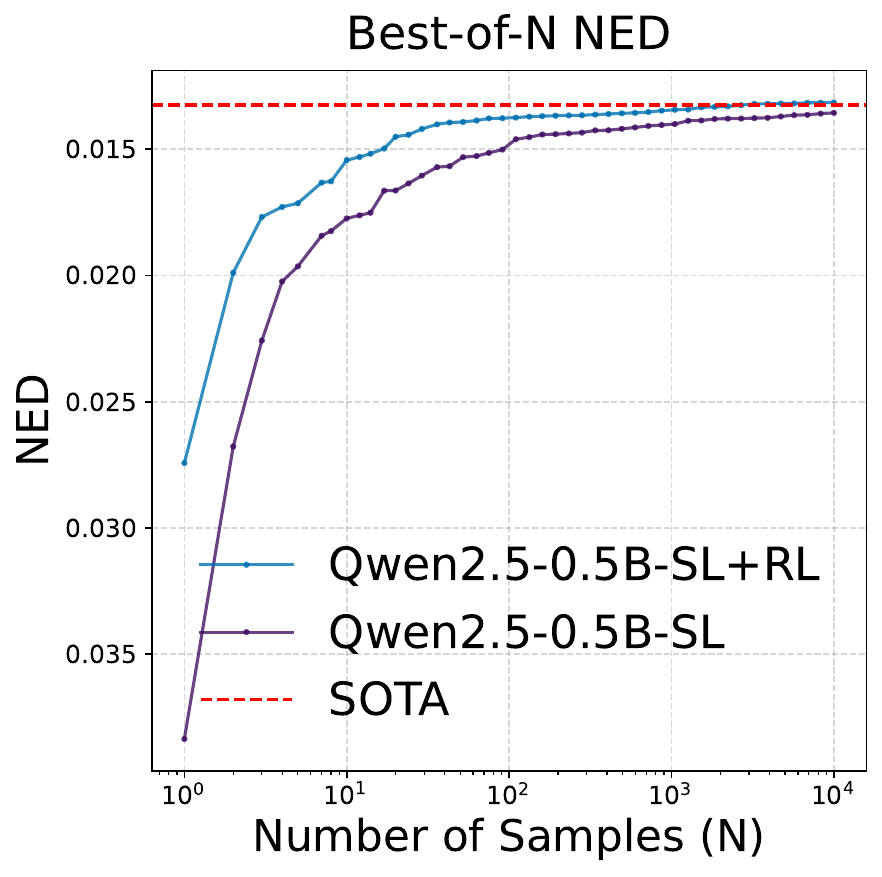}
    &
    \includegraphics[width=0.3\linewidth]{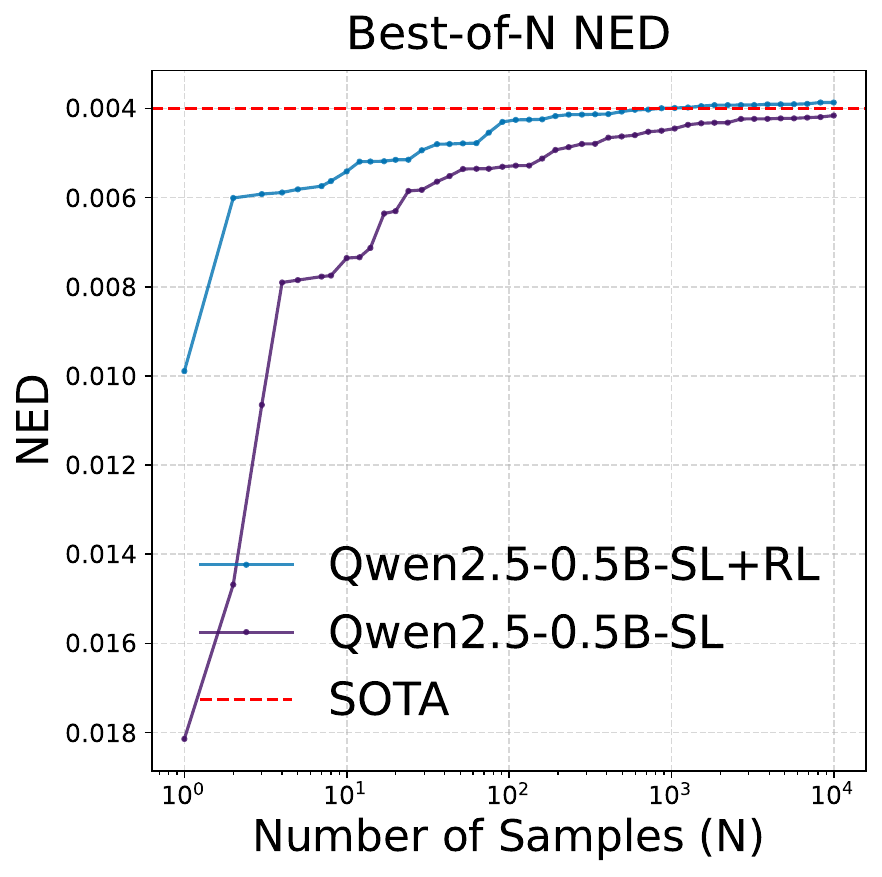}
    \\[-0.2cm]
    (a) Eterna100 & (b) RNAsolo100 & (c) Rfam-Taneda-27
  \end{tabular}
\caption{Best-of-$N$ curves for SL and SL+RL models in terms of Normalized Ensemble Defect (NED) on three data sets. See Fig.~\ref{fig:SL+RL} in the main text for the corresponding Best-of-$N$ curves on Boltzmann probability.}
\label{fig:SI-SL+RL}
\end{figure*}

\begin{figure*}
\centering
  \hspace{-0.2cm}
  \begin{tabular}{ccc}
    \includegraphics[width=0.3\linewidth]{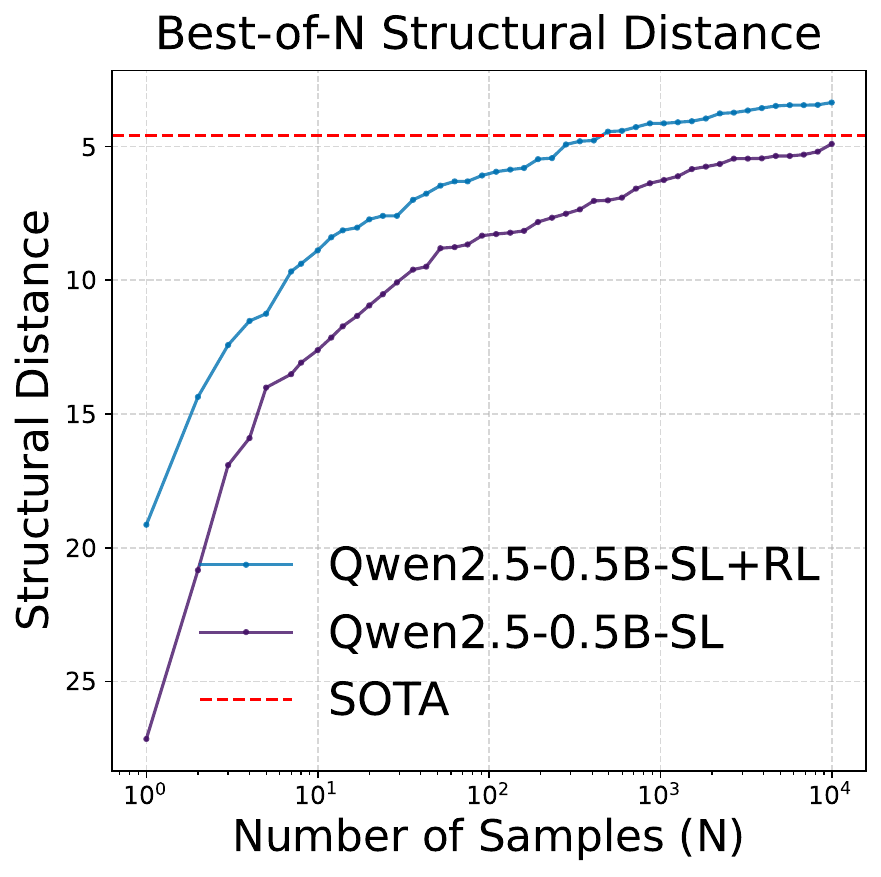}
    &
    \includegraphics[width=0.3\linewidth]{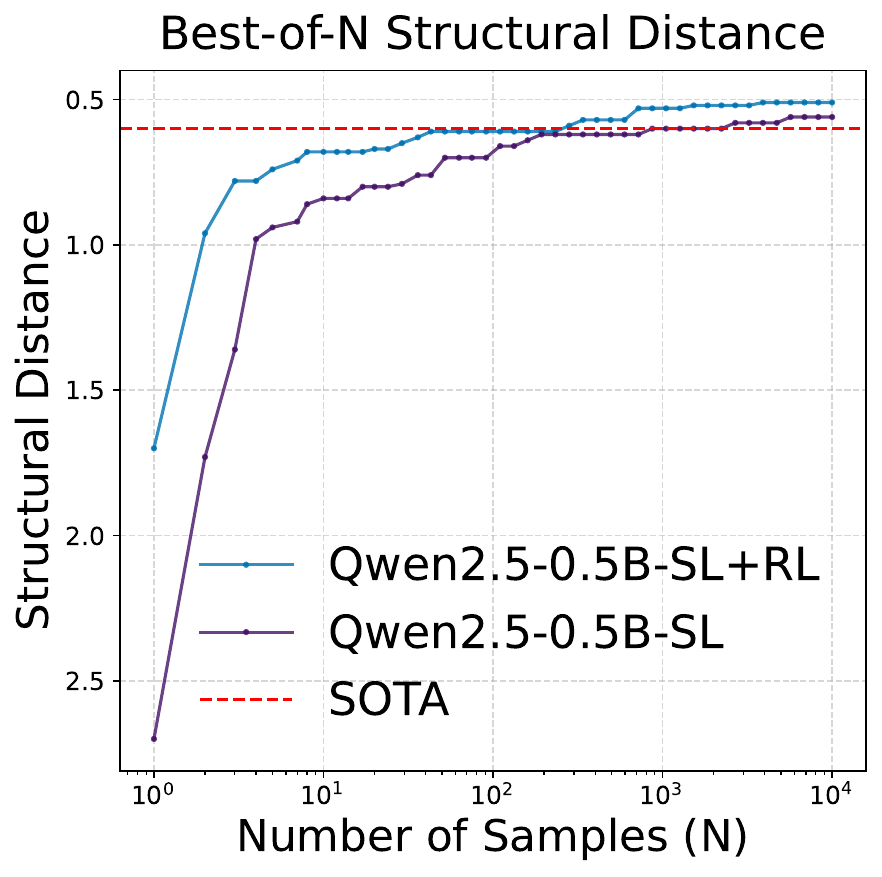}
    &
    \includegraphics[width=0.3\linewidth]{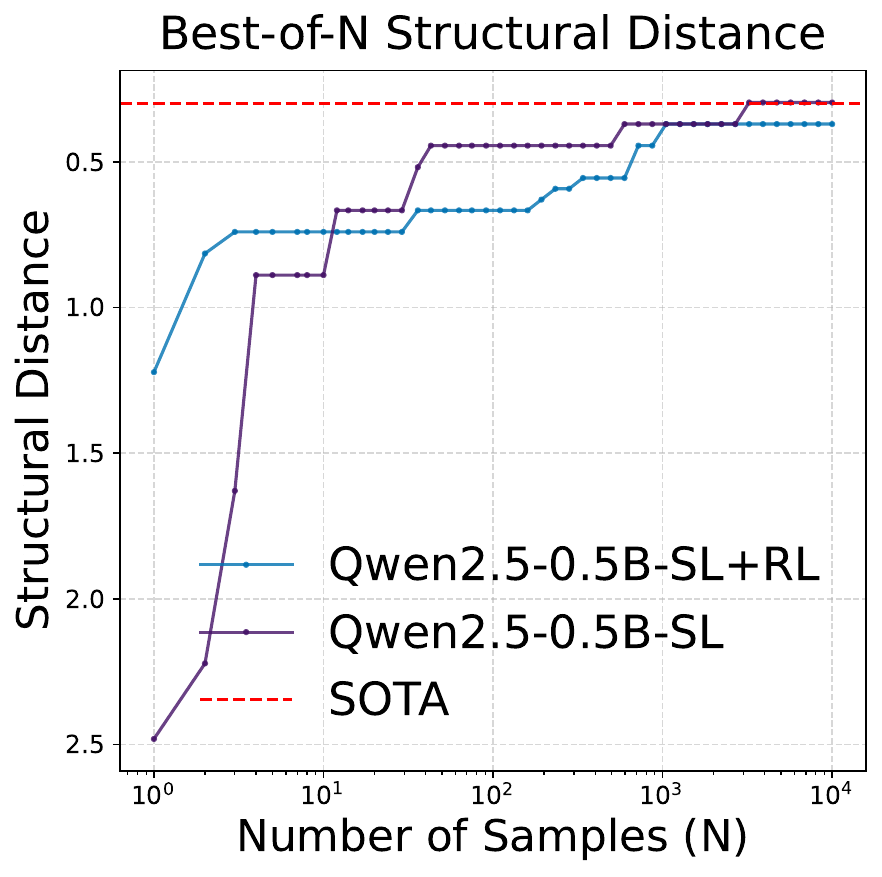}
    \\[-0.2cm]
    (a) Eterna100 & (b) RNAsolo100 & (c) Rfam-Taneda-27
  \end{tabular}
\caption{Best-of-$N$ curves for SL and SL+RL models in terms of Structural Distance on three data sets. See Fig.~\ref{fig:SL+RL} in the main text for the corresponding Best-of-$N$ curves on Boltzmann probability.}
\label{fig:SI-SL+RL}
\end{figure*}

\begin{figure}
    \centering
    \vspace{1cm}
    \includegraphics[width=.5\linewidth]{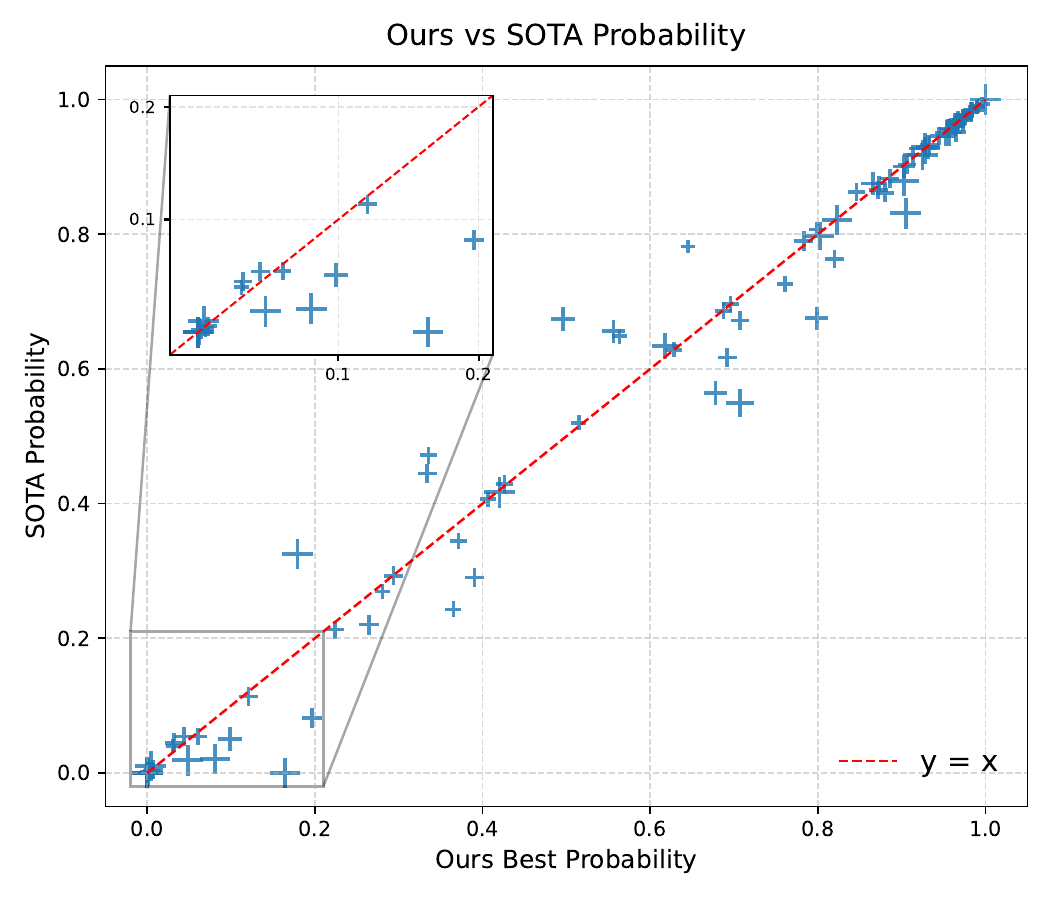}
    \caption{X–Y plot of Boltzmann probability comparing SAMFEO and Ours (SL+RL) model on the Eterna100 test set. Our advantage is most salient on long and hard-to-design puzzles.}
    \label{fig:x-y plot}
\end{figure}

\begin{figure*}
    \centering
    \vspace{-1.5cm}
    \includegraphics[width=.97\textwidth]{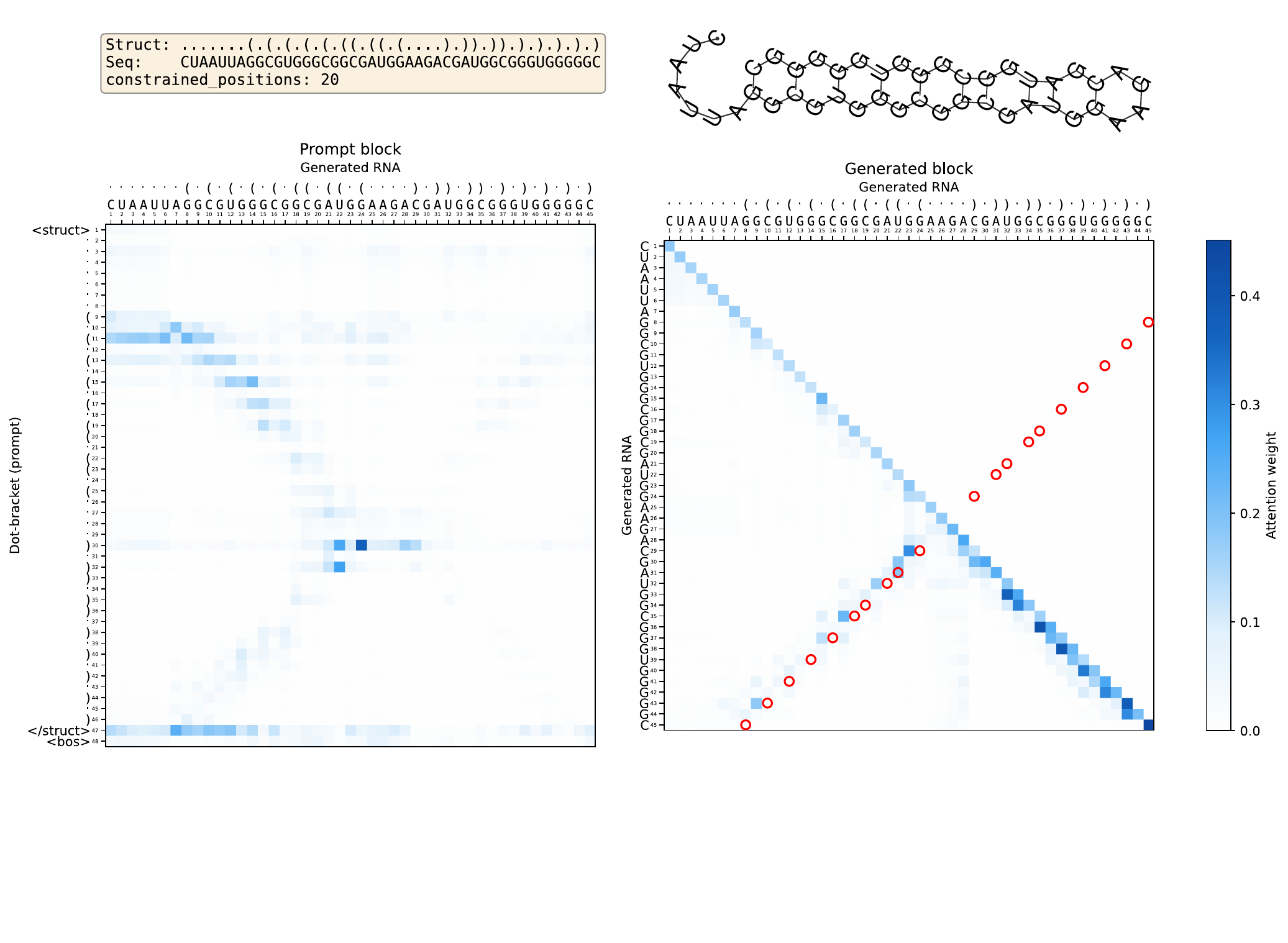}
    \vspace{-2cm}
    \caption{For puzzle \#33 of the Eterna100 test set, the figure visualizes causal self-attention in our decoding. The attention weights shown are averaged over all heads of the last self-attention layer. The left panel shows attention between the prompt (dot-bracket structure) and generated RNA tokens, while the right panel shows attention among generated RNA tokens. The target structure (dot-bracket) and generated RNA sequence appear at the top left, and the corresponding secondary structure is shown at the top right; red circles mark constrained base-pair positions.}
    \label{fig:attention-map}
\end{figure*}

\begin{table}[b!]
\centering
\setlength{\arrayrulewidth}{1pt}
\renewcommand{\arraystretch}{1.3}

\begin{tabular}{l|cc}
\hline
Method 
  & \multicolumn{2}{c}{\# of MFE/uMFE ($\uparrow$)} \\
\cline{2-3}
  & Eterna100
  & Rfam-Taneda-27 \\
\hline
Eastman \emph{et al.} (\cite{eastman+:2018})
  & 60 / 42
  & -- / -- \\
LEARNA (\cite{runge+:2019})
  & 67 / --
  & 23 / -- \\
Meta-LEARNA (\cite{runge+:2019})
  & 68 / --
  & {\bf 24} / -- \\
\hline
Ours (SL+RL)
  & \textbf{75 / 73}
  & {\bf 24} / 24 \\

\hline
\end{tabular}
\vspace{5pt}
\caption{Comparison of RL methods on Eterna100 and Rfam-Taneda-27. The baselines' numbers are taken from those papers and they only reported MFE metrics.
Our work substantially outperforms them on Eterna100.}
\label{tab:SI-RL}
\end{table}

\end{appendices}

\end{document}